\UseRawInputEncoding

\documentclass[a4paper,svgnames,dvipsnames,table, fleqn]{cas-sc}

\usepackage[authoryear,longnamesfirst]{natbib}

\newcommand{\revision}[1]{\textcolor{blue}{#1}}
\usepackage[colorinlistoftodos,prependcaption,textsize=tiny]{todonotes}

\usepackage{tikz}
\usetikzlibrary{shapes.geometric, arrows}

\usepackage{tabularx}
\usepackage{booktabs}
\usepackage{longtable}
\usepackage{array}
\usepackage{amsmath}
\usepackage{graphicx}  % For including images
\usepackage{caption}   % For better captions
\usepackage{subcaption} % For subfigures
\usepackage{float}     % For precise positioning
\usepackage{adjustbox}  % alternative to \resizebox

\usepackage{soul}
\usepackage[normalem]{ulem} % for \sout in text
\usepackage{cancel}         % for \cancel / \xcancel in math

\newcommand{\mycomment}[1]{}
\usepackage{comment}
\usepackage{xcolor}
\usepackage{amssymb}
\usepackage{subcaption}

\usepackage{float}

\def\tsc#1{\csdef{#1}{\textsc{\lowercase{#1}}\xspace}}
\tsc{WGM}
\tsc{QE}
\tsc{EP}
\tsc{PMS}
\tsc{BEC}
\tsc{DE}
\begin{document}
\let\WriteBookmarks\relax
\def\floatpagepagefraction{1}
\def\textpagefraction{.001}

% Short title
\shorttitle{Estimation and Optimization of Ship Fuel Consumption in
Maritime:}

% Short author
%%%\shortauthors{Marijan et~al.}

% Main title of the paper
\title [mode = title]{Estimation and Optimization of Ship Fuel Consumption in
Maritime: Review, Challenges and Future Directions}                      
% Title footnote mark
% eg: \tnotemark[1]
%\tnotemark[1,2]

% Title footnote 1.
% eg: \tnotetext[1]{Title footnote text}
% \tnotetext[<tnote number>]{<tnote text>} 
%\tnotetext[1]{This document is the results of the research
%   project funded by the National Science Foundation.}

%\tnotetext[2]{The second title footnote which is a longer text matter
%   to fill through the whole text width and overflow into
%   another line in the footnotes area of the first page.}

% First author
%
% Options: Use if required
% eg: \author[1,3]{Author Name}[type=editor,
%       style=chinese,
%       auid=000,
%       bioid=1,
%       prefix=Sir,
%       orcid=0000-0000-0000-0000,
%       facebook=<facebook id>,
%       twitter=<twitter id>,
%       linkedin=<linkedin id>,
%       gplus=<gplus id>]
\author[1]{Dusica Marijan}[type=editor,
                        %auid=000,
                        %bioid=1,
                        %orcid=0000-0001-7511-2910]
                        ]

% Corresponding author indication
\cormark[1]

% Footnote of the first author
\fnmark[1]

% Email id of the first author
\ead{dusica@simula.no}

% URL of the first author
\ead[url]{www.simula.no, dusica@simula.no}

%  Credit authorship
%%%\credit{Conceptualization of this study, Methodology, Software}

% Address/affiliation
\affiliation[1]{organization={Simula Research Laboratory},
    addressline={Kristian Augusts 23}, 
    city={Oslo},
    % citysep={}, % Uncomment if no comma needed between city and postcode
    postcode={0164}, 
    % state={},
    country={Norway}}

% Second author
\author[1]{Hamza Haruna Mohammed}[]

% Third author
\author[1]{Bakht Zaman}[%
   %role=Co-ordinator,
   %suffix=BZ,
   ]
%\fnmark[2]
%\ead{cvr3@sayahna.org}
%\ead[URL]{www.sayahna.org}

%%%\credit{Data curation, Writing - Original draft preparation}

% Address/affiliation

% Fourth author
%\author%
%[2]
%\ead{rishi@stmdocs.in}
%\ead[URL]{www.stmdocs.in}

\begin{comment}

% Corresponding author text
\cortext[cor1]{Corresponding author}
\cortext[cor2]{Principal corresponding author}

% Footnote text
\fntext[fn1]{This is the first author footnote. but is common to third
  author as well.}
\fntext[fn2]{Another author footnote, this is a very long footnote and
  it should be a really long footnote. But this footnote is not yet
  sufficiently long enough to make two lines of footnote text.}

% For a title note without a number/mark
\nonumnote{This note has no numbers. In this work we demonstrate $a_b$
  the formation Y\_1 of a new type of polariton on the interface
  between a cuprous oxide slab and a polystyrene micro-sphere placed
  on the slab.
  }

\end{comment}

% Here goes the abstract
\begin{abstract}
To reduce carbon emissions and minimize shipping costs, improving the fuel efficiency of ships is crucial. Various measures are taken to reduce the total fuel consumption of ships, including optimizing vessel parameters and selecting routes with the lowest fuel consumption. Different estimation methods are proposed for predicting fuel consumption, while various optimization methods are proposed to minimize fuel oil consumption. This paper provides a comprehensive review of methods for estimating and optimizing fuel oil consumption in maritime transport. Our novel contributions include categorizing fuel oil consumption \& estimation methods into physics-based, machine-learning, and hybrid models, exploring their strengths and limitations. Furthermore, we highlight the importance of data fusion techniques, which combine AIS, onboard sensors, and meteorological data to enhance accuracy. We make the first attempt to discuss the emerging role of Explainable AI in enhancing model transparency for decision-making. Uniquely, key challenges, including data quality, availability, and the need for real-time optimization, are identified, and future research directions are proposed to address these gaps, with a focus on hybrid models, real-time optimization, and the standardization of datasets. 
\end{abstract}

% Keywords
% Each keyword is separated by \sep

\maketitle

\section{Introduction}
Maritime commercial transport is crucial for global supply chains and international commerce, with over 75\% of goods transported by sea routes \cite{FERRARI2023100985}. However, maritime transport significantly contributes to greenhouse gas (GHG) emissions, leading to pollution and global warming. Therefore, authorities have been making constant efforts to make maritime transport efficient, and strategies are being devised to control greenhouse gas emissions. For example, the International Maritime Organization (IMO) is playing its role by introducing regulations on ship energy efficiency and GHG emissions. According to the 2023 IMO GHG strategy \cite{mepc20232023}, followed by the adoption of mid-term measures at the MEPC in 2025, carbon emissions per vessel transport work are aimed to be reduced by at least 40\% by 2030 compared to 2008. The strategy also aims to reach net-zero GHG emissions by or around 2050, with a mid-term target of at least 70\% reduction by 2040. % 

The primary cost in maritime transport is fuel, with fuel oil consumption (FOC), including all fuel oil consumed on board, accounting for approximately two-thirds of the cruising expenses of a vessel and more than 25\% of the total operating expenses of a ship \cite{gkerekos2019machine}. Thus, reducing FOC would result in reducing GHG emissions and reducing costs. A reduction in FOC can be achieved through multiple methods, including optimizing speed and trim, route optimization, or weather routing. To this end, this Review Article deals with the FOC estimation and FOC optimization techniques.

\subsection{Motivation and Contribution}
Despite the significant advancements in ship fuel consumption estimation and optimization, several gaps remain in the current literature. Existing review papers primarily focus on individual estimation models or optimization techniques, often neglecting the integration of multiple approaches \citet{mylonopoulos2023comprehensive}. While previous studies discuss data-driven and physics-based models separately, there is a limited amount of research on hybrid methodologies that effectively combine both paradigms. Moreover, real-time optimization strategies remain underexplored, which are crucial for operational decision-making in dynamic maritime environments \citet{yan2021data}.

Another critical gap is the lack of standardized datasets for fuel consumption estimation models. The maritime industry relies on diverse data sources, such as Automatic Identification System (AIS) data, onboard sensor readings, and meteorological records \cite{li2024dapnet}. However, the fusion of these data sources for improved estimation accuracy has not been thoroughly investigated \cite{zhu2021modeling}. Additionally, while Explainable AI (XAI) techniques are gaining traction in various domains, their application in maritime fuel consumption remains limited, making it difficult for stakeholders to trust AI-driven decision-making processes \cite{wang2023innovative} \cite{ma2023interpretable}.

Reviews are available on fuel consumption estimation/prediction models for maritime ships such as \citet{fan_review_2022, wang_comprehensive_2022} and fuel optimization models, e.g., \citet{mylonopoulos2023comprehensive}. Moreover, the paper by \citet{barreiro_review_2022} reviews research articles related to the energy efficiency of ships. It also reviews several route and trim optimization methods to improve the energy efficiency of ships.  However, a comprehensive review paper on vessel fuel consumption estimation and optimization models is lacking in the literature. A related review paper by \cite{yan_data_2021} was published in 2021, however, it does not include recent work on fuel consumption estimation and optimization up to 2025.
The novel perspectives and insights presented in this review paper are as follows.
\begin{itemize}
    \item Different data sources relevant to the estimation of FOC are reviewed and discussed in detail. Moreover, a unique aspect of the present review is that the fusion of different data sources for FOC is reviewed and discussed. The paper presents a detailed discussion of the data fusion techniques and a table related to data fusion.  
    \item The nature of the FOC estimation algorithms (batch/online) in the form of the capability to process streaming data is reviewed.
    \item Optimization algorithms for FOC of ships are categorized into several types based on various criteria.
    \item Explainable AI techniques are reviewed in the context of maritime fuel consumption.
    \item The limitations of the available research paper are listed, the challenges of FOC estimation and optimization are discussed, and future directions for FOC estimation and optimization are detailed.  
\end{itemize}
A detailed comparison of the present review paper and other review studies on estimating and optimizing fuel oil consumption in the maritime sector is presented in Table \ref{tab:comparison}.

\begin{table}[h]
    \centering
    \caption{Comparison of relevant survey papers with our paper highlighting the novel contributions of our work (\textbf{FOC Est.}: FOC Estimation, \textbf{FOC Opt.}: FOC Optimization, \textbf{FOC-XAI}: FOC Explainable AI)
    }
    \renewcommand{\arraystretch}{1.5}
    \begin{tabular}{|p{3.5cm}|c|c|c|c|c|c|}
        \rowcolor{gray!20}
        \hline
        \textbf{Papers} & \textbf{FOC Est.} & \textbf{FOC Opt.} & \textbf{FOC-XAI} & \textbf{Feature Discussion} & \textbf{Challenges} & \textbf{Data Fusion} \\ \hline
         \textbf{\cite{fan_review_2022}} & \checkmark & $\times$ & $\times$ & \checkmark & $\times$ & $\times$ \\ \hline
        \textbf{\citet{wang_comprehensive_2022}} & \checkmark & $\times$ & $\times$ & $\times$ & $\times$ & $\times$ \\ \hline
        \textbf{\citet{barreiro_review_2022}} & $\times$ & $\checkmark$ & $\times$ & $\times$ & $\checkmark$ & $\times$ \\ \hline
       \textbf{\citet{mylonopoulos2023comprehensive}} & \checkmark & \checkmark & $\times$ & $\times$ & $\times$ & $\times$ \\ \hline
        \textbf{\citet{yan2021data}} & \checkmark & \checkmark & $\times$ & \checkmark & \checkmark & $\times$ \\ \hline
        \textbf{Our work} & \checkmark & \checkmark & \checkmark & \checkmark & \checkmark & \checkmark \\ \hline
    \end{tabular}

    \label{tab:comparison}
\end{table}

\subsection{Literature Search Protocol
}
We used the following search terms to identify articles related to ship fuel consumption estimation and optimization: "fuel consumption", "fuel optimization", "machine learning models in maritime", "physics-based models in maritime", "hybrid models in maritime", "maritime data fusion", "maritime explainable AI". The search was performed using Scopus, ACM, IEEEXplore, and Google Scholar databases. Initial search returned 1305 articles. To refine the search results, we read the titles and abstracts of these papers and filtered them to select only those related to the estimation and optimization of ship fuel consumption in maritime. If it was not clear whether the paper is relevant from reading the title and abstract only, we read the whole paper. We included both review articles and empirical papers reporting technical work. The filtering resulted in 115 papers. Afterwards, we searched for additional relevant papers by analyzing the references of those selected papers. Finally, we obtained 140 papers used for writing this review article. 

The remainder of the paper is organized as follows. Section 2 presents an overview of the data required for fuel consumption. Section 3 addresses the models used to estimate the fuel consumption of ships. Section 4 presents an overview of the different categories of models for predicting fuel consumption. Section 5 discusses challenges, limitations, and future directions for FOC estimation and optimization. Section 6 presents the conclusion of the paper. Finally, the list of abbreviations used in the paper is presented in Table \ref{tab:abbreviations}.

\begin{longtable}{ll}
\caption{List of abbreviations used in the paper}\label{tab:abbreviations}\\
\hline
\textbf{Nomenclature} & \textbf{Meaning} \\
\hline
\endfirsthead

\hline
\textbf{Nomenclature} & \textbf{Meaning} \\
\hline
\endhead

AIS & Automatic Identification System \\
AES & All-electric ship \\
ANN & Artificial Neural Network \\
BBM & Black Box Model \\
COG & Course Over Ground \\
CMEMS & Copernicus Marine Environment Monitoring Service \\
DCS & Data Collection System \\
DT & Decision Trees \\
ECA & Emission Control Areas \\
ECMWF & European Centre for Medium-Range Weather Forecasts \\
EEMD & Ensemble Empirical Mode Decomposition \\
EMSA & European Maritime Safety Agency \\
EMO & Evolutionary Multi-objective Optimization \\
ETA & Estimated Time of Arrival \\
EU & European Union \\
FOC & Fuel Oil Consumption \\
GBM & Grey Box Model \\
GHG & Greenhouse Gas \\
IMO & International Maritime Organization \\
LSTM & Long Short-term Memory \\
MAE & Mean Absolute Error \\
ME & Main Engine \\
MEPC & Marine Environment Protection Committee \\
MIO & Mixed-integer quadratic optimization \\
MLR & Multiple Linear Regression \\
MRV & Measurements, Reporting, and Verification \\
MOEA & Multi-objective Evolutionary Algorithm \\
MT & Metric Ton \\
NMEFC & National Marine Environmental Forecasting Center \\
NR & Noon Report \\
NOAA & National Oceanic and Atmospheric Administration \\
PI-NN & Physics-informed Neural Network \\
PSO & Particle Swarm Optimization \\
RF & Random Forests \\
RPM & Revolutions Per Minute \\
RMSE & Root Mean Square Error \\
SFOC & Specific Fuel Oil Consumption \\
S-GBM & Serial Grey Box Modeling \\
SHAP & SHAPley Additive exPlanations \\
SVMR & Support Vector Machine Regression \\
SOG & Speed Over Ground \\
WBM & White Box Model \\
XAI & Explainable Artificial Intelligence \\
\hline
\end{longtable}
\section{Data Required for Fuel Consumption Estimation}
\begin{figure}[!ht]
    \centering

    % First row
    \begin{subfigure}[t]{0.492\textwidth}
        \centering
        \includegraphics[width=\linewidth]{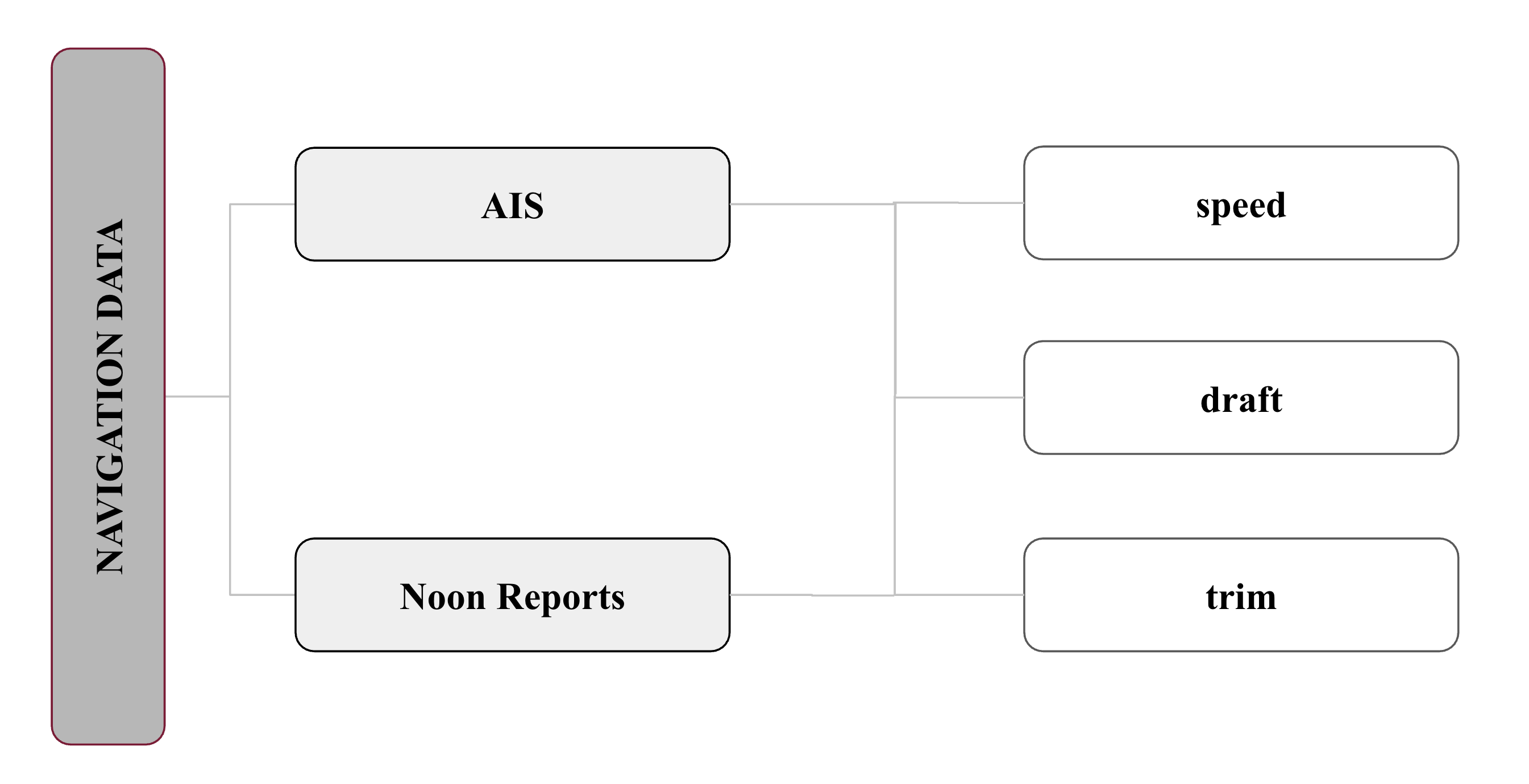}
        \caption{\textbf{Navigation Data: }Derived from AIS and Noon Reports, this includes parameters such as vessel speed, draft, and trim, which are essential for understanding ship movement and loading conditions.}
        \label{fig:nav_data}
    \end{subfigure}
    \hfill
    \begin{subfigure}[t]{0.492\textwidth}
        \centering
        \includegraphics[width=\linewidth]{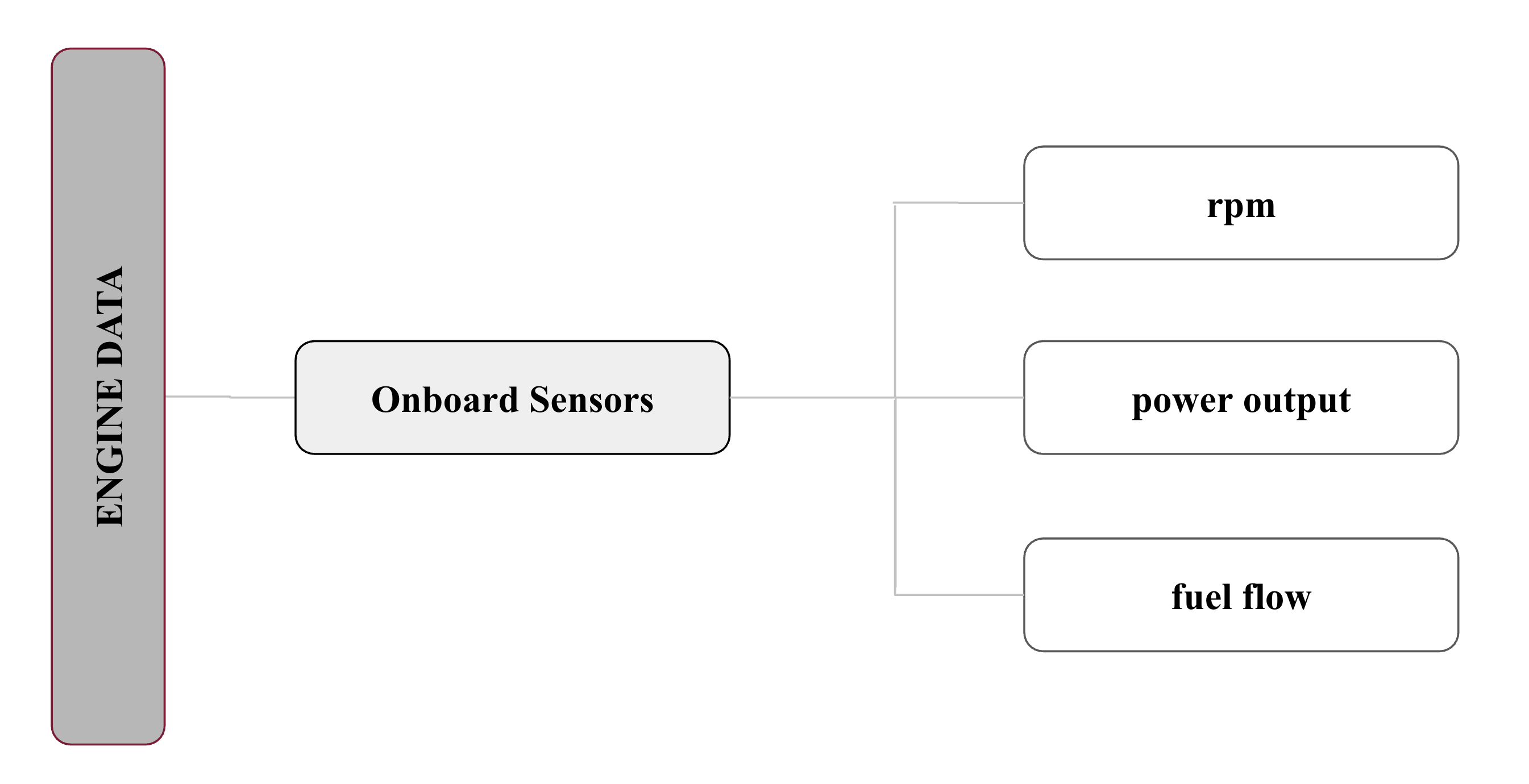}
        \caption{\textbf{Engine Data: }Collected via onboard sensors, capturing real-time parameters such as RPM, power output, and fuel flow to monitor engine performance.}
        \label{fig:engine_data}
    \end{subfigure}

    \vspace{0.5em} % spacing between rows

    % Second row
    \begin{subfigure}[t]{0.494\textwidth}
        \centering
        \includegraphics[width=\linewidth]{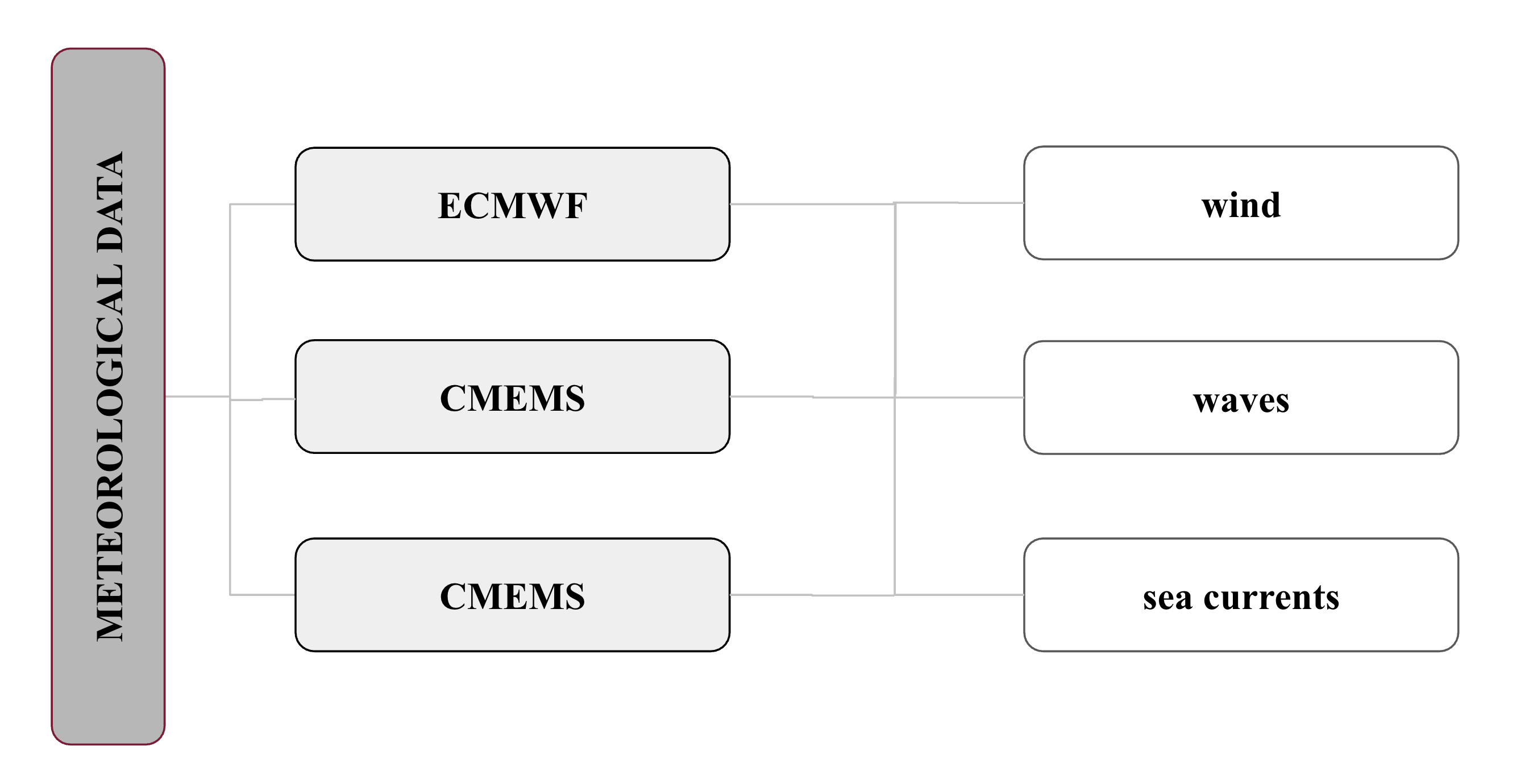}
        \caption{\textbf{Meteorological Data: }Sources include ECMWF and CMEMS, providing data on wind, waves, and sea currents to evaluate environmental conditions affecting vessel performance.}
        \label{fig:met_data}
    \end{subfigure}
    \hfill
    \begin{subfigure}[t]{0.494\textwidth}
        \centering
        \includegraphics[width=\linewidth]{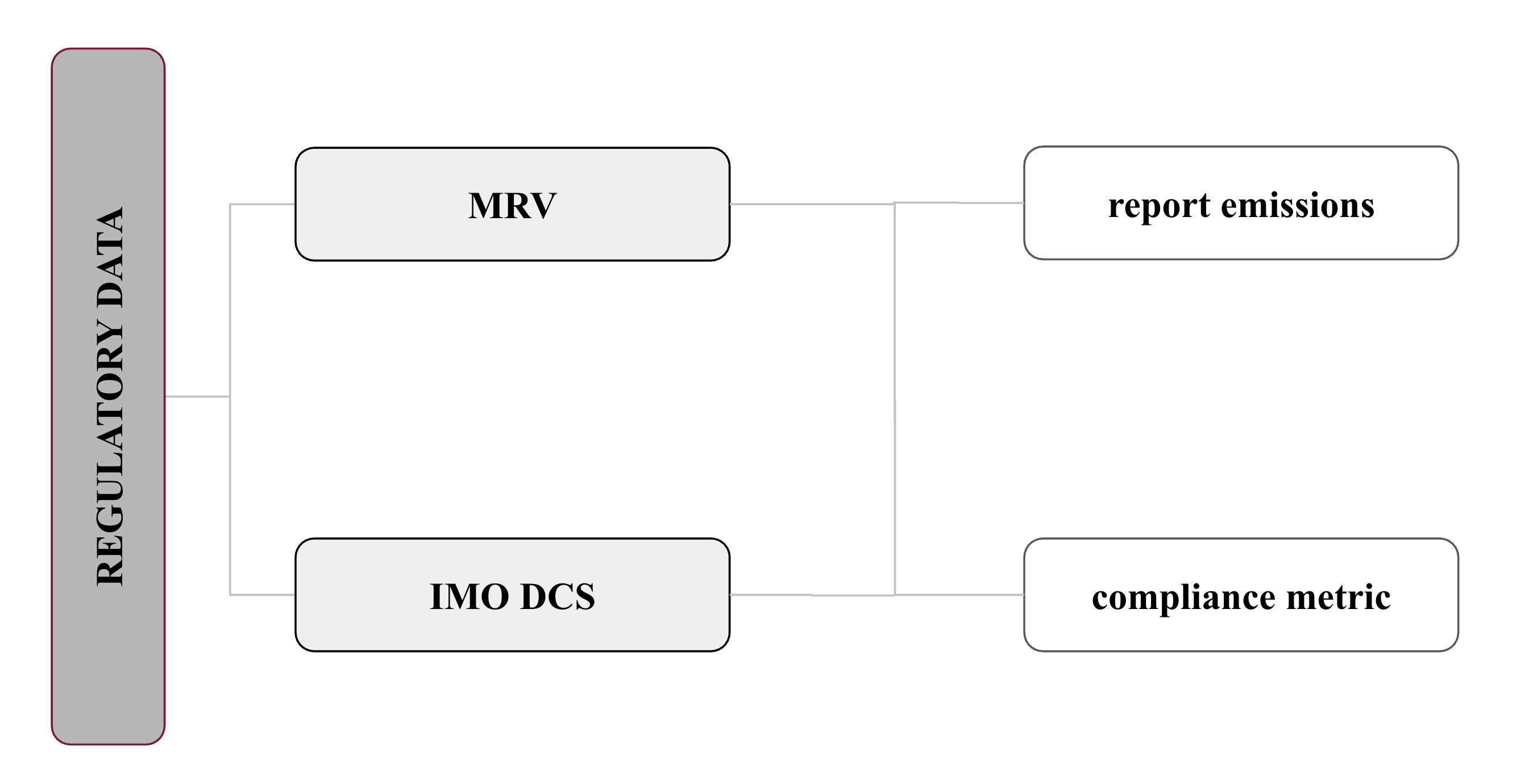}
        \caption{\textbf{Regulatory Data: }Includes MRV and IMO DCS systems, used for emission reporting and calculating compliance metrics by international standards.}
        \label{fig:reg_data}
    \end{subfigure}

    \caption{FOC Estimation Data Sources: Overview of the key data types used in estimating FOC, incorporating navigational, engine, environmental, and regulatory inputs.}
    \label{fig:foc_data_source}
\end{figure}

Data in various forms are observed and collected in maritime ships for monitoring, maintenance, historical records, compliance with regulations, analysis, estimation, and optimization purposes, as illustrated in Table \ref{tab:foc_data_sources}. The data sources used for FOC estimation differ considerably in their resolution and accessibility, as summarized in Table~\ref{tab:data_sources_comparison}. Publicly available data such as AIS and meteorological records facilitate large-scale studies, whereas onboard sensor and MRV data, although more precise, are often restricted by proprietary or regulatory constraints. 
This heterogeneity emphasizes the need for standardized, open benchmarking datasets in maritime analytics. Most of the data is recorded automatically by the systems installed on the ships. However, some of the data are manually recorded by the crew. Since the present paper is related to FOC estimation and optimization, we only discuss the data sources relevant to the fuel consumption of ships as shown in Figure \ref{fig:foc_data_source}. Various data sources related to FOC in maritime ships are detailed in the following:

\begin{table}[!ht]
    \centering
     \caption{Data sources, their corresponding features, and example sources relevant for FOC estimation.}
    \label{tab:foc_data_sources}
    \renewcommand{\arraystretch}{1.8} % Adjust row height for better readability
    \begin{tabularx}{\textwidth}{p{4.8cm} p{4.8cm} p{5.6cm} }
     \rowcolor{gray!20}
        \toprule
        \textbf{Data Type} & \textbf{Example Sources} & \textbf{Key Features} \\
        \midrule
        \textbf{Navigation Data} & AIS, Noon Reports & speed, draft, trim \\
        \textbf{Engine Data} & Onboard Sensors & rpm, power output, fuel flow \\
        \textbf{Meteorological Data} & ECMWF, NOAA, CMEMS & wind, waves, sea currents \\
        \textbf{Regulatory Data} & MRV, IMO DCS & reported emissions, compliance metrics \\
        \bottomrule
    \end{tabularx}
   
    \label{tab:foc_data_sources}
\end{table}

    \subsection{Noon Reports}
    A noon report (NR) is prepared daily (generally at local noon) and manually entered by the ship's crew, which contains data and information on the ship's state during a voyage. The noon reports contain the current geographical location, speed, course, distance traveled since the last report, loading condition, average rpm of the main engine, shaft power, as well as information about the weather and sea conditions such as wind, waves, currents, sea temperature, etc. \cite{yan2021emerging, du_two-phase_2019}. Essentially, noon reports contain the necessary information to estimate ships' fuel consumption. Noon reports are a common data source in FOC estimation, particularly for daily fuel consumption analysis \cite{icsikli2020estimating}. NR data can be helpful in long voyages that last several days; however, for short voyages and short-term fuel consumption prediction, NRs are not useful \cite{agand_fuel_2023,bayraktar2024marine}. 
    
    \subsection{Onboard Sensors Data}
    In modern ships, the standard data source for ship operators is data from onboard sensors. These sensors are installed to measure relevant quantities related to the environment, engine, position, and fuel consumption in various parts of the ship. Related to FOC, relevant onboard sensors include those for measuring RPM, engine room temperature, engine power, fuel flow, shaft power meters, wind speed, water depth, latitude, longitude, shaft speed, shaft power, and ship speed.
    The onboard sensors collect data at intervals of several seconds. However, these data are proprietary to the ship's owners and are not publicly accessible.
    Several research works only consider data from sensors onboard in fuel consumption models, such as \cite{fan_comprehensive_2024, papandreou2022predicting}.
    
    \subsection{AIS Data}
    AIS is the leading data source in maritime analysis and monitoring for several reasons, including its open-source nature. Each ship equipped with an AIS transceiver broadcasts static (ship name, type, size) and dynamic information (latitude, longitude, SOG, COG) about the ship. AIS also includes information related to the voyage, such as the destination name, draft, and estimated time of arrival (ETA). The transmission rate for AIS messages is approximately a few seconds for a fast-moving ship; however, the recorded AIS data has a frequency of minutes, on average. Recently, a notable increase has been observed in the use of AIS data to train machine learning models for anomaly detection, route extraction, traffic prediction, and collision avoidance systems. AIS data are also mainly used for fuel consumption estimation, along with other data sources. For instance, \citet{weng_ship_2020} uses AIS data to estimate ship emissions, while other works, such as \cite{du_data_2022, yuan_prediction_2021, yan2024improving}, combine AIS data with other sources of data for FOC estimation.
    
    \subsection{Meteorological Data}
    The meteorological and sea data are also for fuel consumption and vessel safety, in general. Weather and sea conditions affect the ships' fuel consumption; therefore, these conditions are used in the estimation and prediction of the FOC of the ships. The meteorological and sea data variables include wind speed, wave height, sea currents, seawater temperature, sea depth, seawater density, and others. Relevant data related to weather and sea conditions, along with their forecasts, are available on websites such as the European Centre for Medium-Range Weather Forecasts (ECMWF), windy.com, the Copernicus Marine Environment Monitoring Service (CMEMS), the National Oceanic and Atmospheric Administration (NOAA), and the National Marine Environmental Forecasting Center (NMEFC).
    
    \subsection{Measurements, Reporting, and Verification (MRV)} 
    The European Union (EU) introduced a monitoring, reporting, and verification (MRV) system in 2018 for large vessels that enter, leave, or operate in EU ports to reduce greenhouse gas emissions \cite{luo2023after}. MRV provides an opportunity to gather information about the ships' fuel consumption, carbon emissions, and energy efficiency. It is worth noting that the data collected by the MRV system is publicly accessible through the European Maritime Safety Agency (EMSA). It has already been used as a benchmark for comparison in research work such as \cite{kim_modelling_2023}. Some works use only MRV data to train machine learning models for predicting FOC, e.g., \cite{yan2023analysis}. Other works that use MRV data and analyze FOC include \cite{ren2022container} and \cite{doundoulakis2022comparative}. 
    In \citet{heikkila2024effect}, openly available MRV data and other data sources are used to analyze the fuel consumption of ice-classed ships.

    After discussing all data sources related to FOC for ships in maritime, an overview of the data types and their data sources is presented in Table \ref{tab:datasources}.

\begin{table}[h]
    \centering
      \caption{Data sources and their corresponding features related to engine performance, navigation, metocean conditions, FOC, and GHG emissions.}
    \renewcommand{\arraystretch}{1.5}
    \begin{tabular}{|p{1.3cm}|p{2.8cm}|p{2.0cm}|p{4.3cm}|p{3.8cm}|}
        \rowcolor{gray!20}
        \hline
        \textbf{Source} & \textbf{Engine} & \textbf{Navigation} & \textbf{Metocean} & \textbf{FOC and GHG Emissions} \\
        \hline
        \textbf{AIS} &  & SOG, COG, draft &  &  \\
        \hline
        \textbf{Noon Reports} & average propeller rpm, average engine rpm & draft, trim, average SOG, COG & wind magnitude and direction, sea conditions and sea swell & Fuel oil consumption (FOC), daily consumption \\
        \hline
        \textbf{Onboard Sensors} & rpm, engine power, torque, engine temperature & SOG, COG & wind magnitude and direction, sea waves & FOC (if fuel sensors are installed) \\
        \hline
        \textbf{Online Sources} &  &  & sea water temperature, wind direction, wind magnitude, wave direction, wave height, sea current direction, sea current speed &  \\
        \hline
        \textbf{MRV} &  &  &  & FOC, CO$_2$ emissions \\
        \hline
    \end{tabular}

    \label{tab:datasources}
\end{table}

\begin{table}[!ht]
\centering
\caption{Summary and comparison of key data sources relevant to FOC estimation.}
\renewcommand{\arraystretch}{1.8}
\rowcolors{2}{gray!10}{white}
\begin{tabularx}{\textwidth}{p{2.3cm} p{2.5cm} p{2.8cm} p{2.8cm} X}
\toprule
\rowcolor{gray!30}
\textbf{Data Source} & \textbf{Example Providers} & \textbf{Typical Features} & \textbf{Resolution / Access} & \textbf{Advantages and Limitations} \\
\midrule

\textbf{AIS} 
& MarineTraffic, EMSA 
& SOG, COG, position, draft, ETA 
& Sec–min; public
& Standardized and global coverage; lacks direct engine/fuel data; possible signal gaps and spoofing; timestamps may need resampling. \\

\textbf{Noon Reports} 
& Ship logbooks 
& Avg. speed, RPM, draft, trim, FOC 
& Daily; internal 
& Contains direct daily FOC; coarse granularity and subject to manual entry noise; limited for short-term/real-time analysis. \\

\textbf{Onboard Sensors} 
& ECU, shaft/fuel meters 
& RPM, torque, shaft power, fuel flow, temp/pressure 
& Seconds; proprietary 
& High fidelity and suitable for real-time modeling; access restricted by confidentiality; sensor drift and calibration needed. \\

\textbf{Online Sources} 
& ECMWF, NOAA, CMEMS 
& Wind, wave, current, SST, pressure 
& Hour–daily; open 
& Essential for weather routing and exogenous features; spatial/temporal mismatch with vessel tracks; interpolation and co-location required. \\

\textbf{MRV} 
& EMSA (EU MRV) 
& Annual FOC, CO\textsubscript{2}, vessel data 
& Annual; public (EU) 
& Regulatory-grade, standardized aggregates; low temporal resolution limits operational analytics; useful as benchmarking/validation. \\

\bottomrule
\end{tabularx}
\label{tab:data_sources_comparison}
\end{table}

\section{Models for Fuel Oil Consumption Estimation}
There are three types of models available in the literature for estimating the fuel oil consumption of maritime ships, i) physics-based models, also known as white box models (WBM), ii) data-driven models, also known as black box models (BBM), and iii) hybrid models, which combine WBMs and BBMs, shown in Figure \ref{fig:taxonomy_of_foc_estimations}. We discuss these three types of models in detail in the following:

\begin{figure*}[!ht]
    \centering
    \includegraphics[scale=0.185]{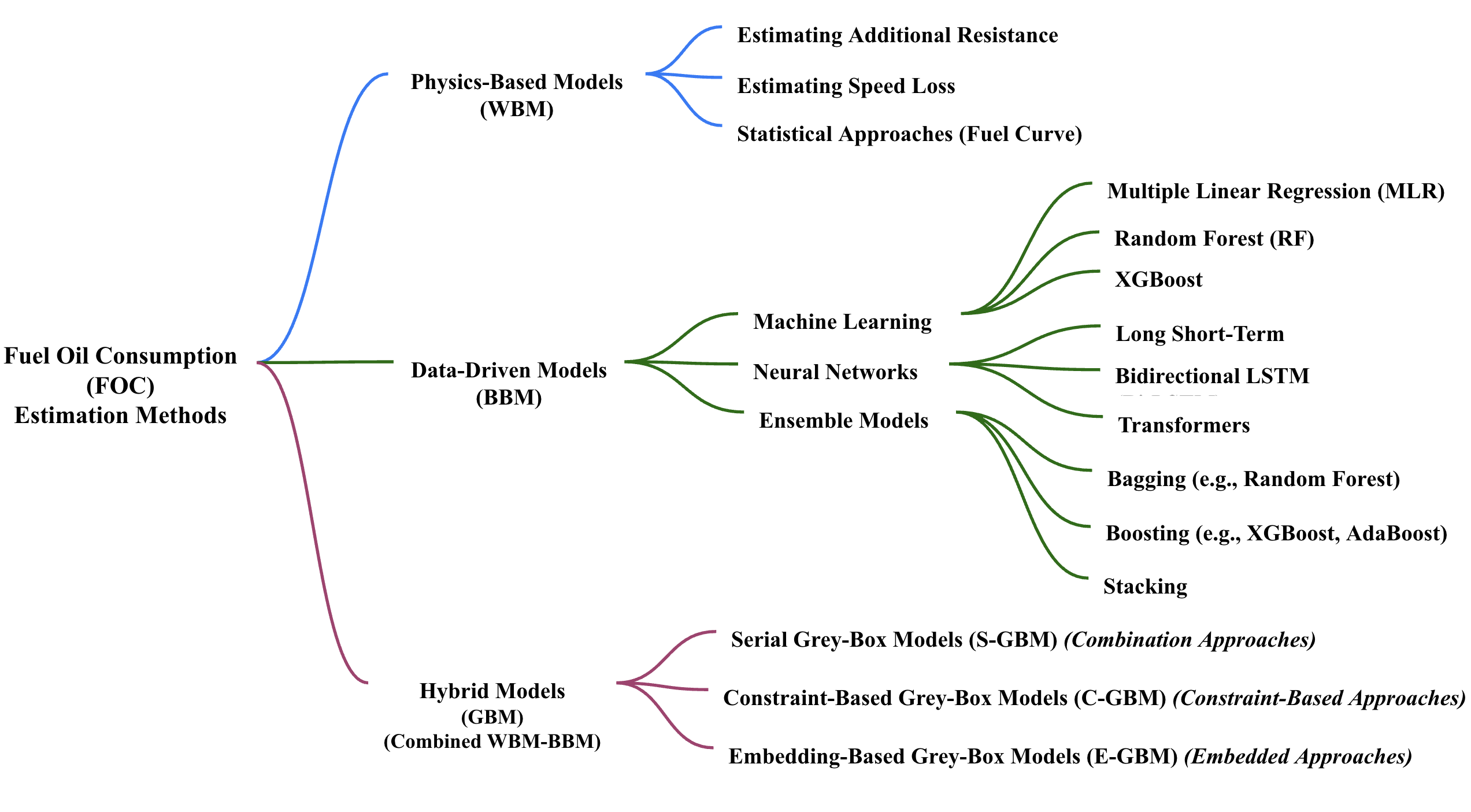}
    \caption{Taxonomy of FOC estimation models}
    \label{fig:taxonomy_of_foc_estimations}
\end{figure*}

\subsection{Physics-based Models}
Physics-based methods involve calculating the total resistance offered to the ship based on the principles of physics, aerodynamics, and hydrodynamics. The total ship resistance is the summation of calm water, wind, waves, marine fouling, and shallow water resistance. The power required for the main engine to drive the ship is proportional to the product of the total resistance and the speed of the ship. Fuel consumption is computed from the instantaneous power and specific fuel oil consumption (SFOC) for a given loading condition. Physics-based methods do not require extensive historical data; however, they do require some information about the ships' components and their parameters for calculations. This may involve information that is not publicly available. The primary papers on physics-based models for estimating FOC include \cite{guo_combined_2022} and \cite{kim_modelling_2023}.

Physics-based models for estimating FOC can be divided into three categories. The first type involves estimating the additional resistance caused by environmental factors. The second type estimates the speed loss due to environmental factors. The third type derives the relationship between the FOC/power and the speed of the vessels. We discuss these models in the following subsections.

\subsubsection{Estimating Additional Resistance}
\begin{figure*}[ht]
    \centering
    \includegraphics[scale=0.20]{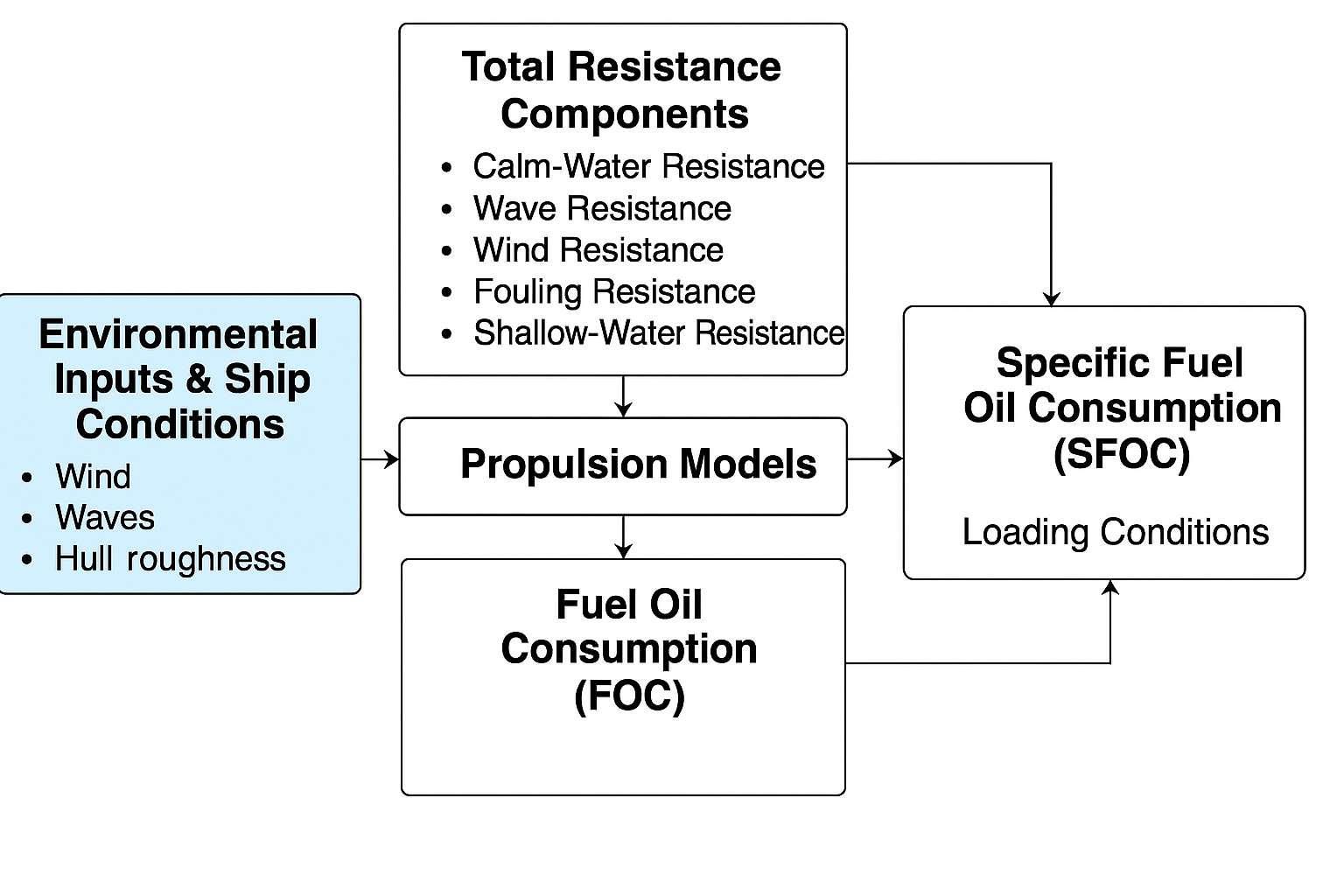}
    \caption{Schematic representation of a resistance-based model for FOC estimation. The process begins with environmental and ship condition inputs (e.g., wind, waves, current, draft, and hull roughness) to compute the total resistance components (calm-water, wave, wind, fouling, and shallow-water resistance). These are then translated into effective engine power requirements through propulsion models. Finally, fuel oil consumption is estimated using SFOC and loading conditions.}
    \label{fig:resistance_based_foc}
\end{figure*}

One method for computing the FOC using physics-based models involves calculating the added resistance encountered by the ship due to waves, wind, current, marine fouling, and calm water, as illustrated in the Figure \ref{fig:resistance_based_foc}. Once the total resistance of the ship is computed, the brake power of the main engine is calculated for a given speed of the ship. Finally, the FOC is computed based on the ship's loading information. 
To compute the total resistance of the given ship, \citet{kim_modelling_2023} proposes a model called the Marine Transport Environmental Assessment Model (MariTEAM), which can determine an appropriate empirical method to estimate the calm water resistance of a ship and calculate the final brake power of the ship. The primary focus of the paper is to calculate the power required under specified conditions. 

In another paper \cite{guo_combined_2022}, a physics-based FOC model is presented, which is an improved version of the VERDE model proposed in \citet{tvete2020modelling}. All added types of resistance are computed via formulas, which also require some parameters related to the ship.  These values were taken from the ship information database. ML is used to impute missing values in the data. Once the total resistance of the ship is computed using the ship propulsion model, given the total resistance and the ship's speed, the adequate power required by the main engine is calculated.  The efficiency of the ship hull, the relative rotation efficiency, and the efficiency of the shaft are discussed. The engine model for ship fuel consumption also considers different fuel types. Machine learning is used for two purposes: 1) to impute missing values and 2) to improve computational efficiency. The authors in \cite{fan_novel_2020} also compute the total resistance offered to the ship, followed by computing the power required, and finally, the FOC is computed.
Similarly, \cite{lu2013voyage} and \cite{zhang2023development} also consider estimating the total resistance of the ship to compute the FOC for a given speed and environmental conditions.

\subsubsection{Estimating Speed Loss}
In this type of physics-based modeling, the speed loss due to environmental and sea factors is calculated, rather than computing the total resistance encountered by the ship. 
The loss of speed due to sea waves is calculated in \cite{lang2021practical} and \cite{lang2020semi} by calculating the added resistance considering different heading angles and different forms of the ship hull, respectively. In addition, the impact of the improved model for speed loss prediction on voyage optimization is also investigated in \citet{lang2021practical}. Moreover, \cite{du2021estimation} also estimates the speed loss due to winds and waves for fuel consumption estimation while finding the optimal route for given constraints. Similarly, \citet{sasa2021speed} presented an algorithm to analyze speed loss due to weather factors. 

\subsubsection{Statistical Approaches Fuel Curve}
The relation between the FOC and the vessel speed is also important, and various approaches have been proposed to derive this relation. For example, in \cite{bialystocki2016estimation}, characteristics such as draft, displacement, wind, sea conditions, current, hull, and propeller roughness are used to predict the FOC of the ship and derive the speed curve by a statistical approach. A statistical approach for the fuel curve is also presented in \cite{kee2018prediction}.

There is also a famous concept known as the "cubic law" of speed and FOC or main engine power; however, it is not relevant after the ship has been in service for a specific time, such as one year, according to \cite{adland2020optimal}. Another approach is to investigate the speed-power relationship for the vessel below the design speed, which involves computing the problem of finding the speed-power exponent, as presented in \citet{berthelsen2021prediction} and \citet{godet2023prediction}. In works similar to \citet{tillig2018analysis}, the fuel consumption of the ship is expressed in terms of the different types of resistances, environmental parameters (air density, seawater densities, water depth, wind speed, etc.), main engine SFOC, parameters related to the structure of the ship, loading conditions (draft) and other efficiency types (hull, propeller, etc.). The FOC is predicted using the values for all these parameters and variables.
    
\subsection{Data-driven Models}
As discussed above, various types of data related to FOC are generated at different stages. Some of these variables and features are not directly related to FOC; however, they can be leveraged to monitor, estimate, and predict the ship's fuel consumption. Data-driven models require a large amount of historical data for training purposes. Once trained, data-driven models can predict the FOC for a given set of features \cite{karagiannidis2021data, fan_comprehensive_2024}. For example, data-driven models for FOC are presented in Table \ref{tab:datadriven-papers}. Table \ref{tab:foc_data_sources_methods} presents a list of papers that consider data-driven models for FOC.

%Models that directly estimate from FOC from the feature. Models indirectly estimating FOC. 
Given the features related to FOC, some of these data-driven models estimate another variable related to fuel consumption. For example, in \cite{kaklis2023enabling}, the authors do not directly estimate the FOC from the given features. Instead, the idea is to estimate the rpm from the observations of velocity measurements. With a model that predicts the rpm, one can calculate an initial estimate of the FOC using SFOC [g/kWh] and some additional information about the ship. The analytical approach for estimating rpm involves constructing piecewise polynomial models over Delaunay Triangulations that divide a suitably defined two-dimensional space of velocity observations. Given the observation of speed, B-splines and neural networks are combined to predict the rpm value. Finally, FOC is a function of rpm and other variables such as propeller diameter, torque coefficient, SFOC, and seawater density. In another example, data-driven methods are employed in \cite{lang2022comparison} to estimate the propulsion power of a ship (which is related to the FOC) using features such as speed, draft, heading, wave, and wind.

There are also approaches available for computing physics-based quantities, such as ship shaft power, using historical real data for a range of ship and sea conditions, employing machine learning methods. For example, \citet{parkes2018physics} deals with the prediction of shaft power using neural networks. Similarly, \citet{lang2022comparison} compares different machine learning algorithms to predict the propulsion power of the ship given the features, including speed through water, mean draft, trim, heading, significant wave height, mean wave period, and wind speed. Note that the fuel consumption is computed based on the power required for the ship. Other works that exploit machine learning for the prediction of power from the shaft of the ship and the calculation of resistance include \cite{laurie2021machine, liang2019prediction, gunecs2023predicting, parkes2019efficient}.

\subsubsection{Methods Used in Data-driven Models}
There are various types of data-driven models; we classify them into the following main categories:\\

\textbf{Classical machine learning:} 
Supervised machine learning algorithms, such as random forests (RF), decision trees (DT), and support vector machine regression (SVMR), are frequently employed for FOC prediction. Moreover, other ML algorithms like multiple linear regression (MLR), XGBoost, Lasso, elastic nets, adaptive boosting, etc., are also used applied in this task   \cite{wang2018predicting, gkerekos2019machine, yan2020development, fan_comprehensive_2024,yuksel_comparative_2023, uyanik_machine_2020, xie_fuel_2023, zhou2023predicting, su2023fuel}. To analyze the uncertainty and resilience of machine learning models in applications, the authors in \cite{nguyen_application-oriented_2023} propose a testing regime that provides insights into models’ behaviors, dependency on different features, and potential vulnerabilities to data uncertainties during the deployment phase.\\

\textbf{Neural networks:}
In recent times, due to significant advances in computer computational power and the availability of vast volumes of data, neural networks have emerged as a key method for estimating fuel consumption. For FOC estimation, \citet{zhang_deep_2024} employs a Bi-LSTM with an attention mechanism, utilizing features such as SOG, draft, trim, main engine (ME) shaft power, ME temperature, and the speed and direction of wind, waves, and sea currents. While a standard LSTM model only considers information from past time frames in the data stream, the Bi-LSTM layer addresses this limitation by incorporating relevant information from past time frames. This is possible as it comprises both forward and backward LSTM sublayers.

Several notable works on FOC estimation using neural networks include \cite{moreira2021neural,shu2024investigation,zhang_deep_2024,bui-duy_utilization_2021,hu_prediction_2019}.
There are also approaches available that predict FOC without using other features, e.g., \citet{chen_short-term_2024}. The paper proposes a BiLSTM network for the short-term prediction of fuel consumption. It utilizes Ensemble Empirical Mode Decomposition (EEMD), a noise-assisted method for decomposing time series. EEMD achieves robustness by adding noise to the signal, which helps obtain more accurate intrinsic mode functions and reduces pattern aliasing. EEMD-LSTM involves applying LSTM to the individual components after decomposition to predict them separately. The predicted values of the components are used to reconstruct the final fuel consumption prediction result. By performing EEMD decomposition on the fuel consumption sequence, the inherent pattern and trend of the distribution can be obtained, which removes the random noise effects. \citet{ilias_multitask_2023} presents a BiLSTMs-based deep learning approach for simultaneously predicting the main and the auxiliary engines.\\

\textbf{Ensemble Models:} To build a model having high robustness, better accuracy, and improved generalization property for fuel consumption prediction, to ensemble various models based on stacking theory \cite{hu_novel_2021}, \cite{hu2022two}. The structure consists of two layers. In the initial layer, three base models are trained, and predictions are made using the training and test sets. This process generates new training and test sets. The subsequent layer features a meta-model based on MLR, which is trained with the new training set and then predicts the new test set to achieve the final results. A Bayesian optimization method of hyperparameters was used to enhance model performance. The advantages of a hybrid model are better generalization ability, the ability to adapt to more complex tasks, the ability to fit nonlinear relationships, and greater robustness.

\subsubsection{Features Related to FOC}
In machine learning-based works for FOC estimation, various types of features are utilized for training the model. It includes both primary (features that are directly observed on the ship) and derived features. In primary features, we have speed, engine rpm, engine temperature, trim, etc. Some of these features are not always used for FOC estimation. For example, engine temperature is not commonly used as a feature in data-driven models; however, it is introduced in \citet{yuan_prediction_2021}, and it is shown that engine temperature affects FOC. Similarly, marine fouling also affects vessels' fuel efficiency as it causes speed loss \cite{coraddu_data-driven_2019}. Therefore, the last date of propeller and hull cleaning during dry docking can be a crucial factor in estimating a ship's fuel consumption. Table \ref{tab:datasources} presents a complete list of characteristics.

The derived features include slip, which is the difference between the actual speed and the theoretical speed of the vessel \cite{yuksel_comparative_2023}.
The propeller slip is the difference between the theoretical distance and the actual distance traveled by the propeller \cite{bayraktar_marine_2024}.
 The loss of speed due to wind, waves, and currents is calculated \cite{hajli_fuel_2024}, and can be used as a feature for FOC in a data-driven model.

\subsubsection{Different Forms of Fuel Consumption} Fuel consumption can be estimated in different forms, for example, metric ton/day or metric ton/hour. Some works predict fuel consumption per distance unit, for example, \cite{hajli_fuel_2024}. The work in \citet{kim2021development} predicts fuel efficiency in tons per nautical mile. 

\subsubsection{Batch or Online Algorithms}
Most data-driven algorithms available to estimate fuel consumption are batch; they need all data before processing. There is a minimum of work that can be done to process the streaming data or retrain the model online. For example, \citet{kaklis_online_2022} can train the model online. Similarly, the prediction of FOC in real time is considered in \cite{yuan_prediction_2021}. The authors in \citet{chi2018framework, chi2015ais} present a framework for real-time monitoring of a ship's energy efficiency.

\begin{table}[h]
    \centering
      \caption{Papers on data fusion of different sources (AIS, noon reports, meteorological, MRV, and onboard sensors data) for FOC estimation.}
     \renewcommand{\arraystretch}{1.5} % Optional: improve row spacing
    \begin{tabular}{|p{8.2cm}|c|c|c|c|c|}
      \rowcolor{gray!20}
        \hline
        \textbf{Papers} & \textbf{Noon Reports} & \textbf{MetOcean} & \textbf{MRV} & \textbf{AIS} & \textbf{Sensors} \\ \hline
        \cite{li_data_2022}, \cite{hajli_fuel_2024}, \cite{luo2023comparison} & \checkmark & \checkmark & $\times$  & $\times$  & $\times$  \\ \hline
        \cite{zhou2023predicting} & $\times$  & \checkmark & $\times$  & $\times$  & \checkmark \\ \hline 
        \cite{du_data_2022,yan2024improving} & \checkmark & \checkmark & $\times$  & \checkmark & $\times$  \\ \hline
        \cite{du_data_2022-1} & $\times$  & \checkmark & & $\times$  & \checkmark  \\ \hline
        \cite{ren2022container}, \cite{wang2023ship} & $\times$  & \checkmark & \checkmark & \checkmark & $\times$  \\ \hline
        \cite{farag2020development,zhang2023research} & $\times$  & \checkmark & $\times$  & \checkmark & \checkmark \\ \hline
        \cite{zhu2021modeling} & \checkmark & $\times$  & $\times$  & $\times$  & \checkmark \\ \hline
        \cite{heikkila2024effect} & $\times$  & \checkmark & \checkmark & \checkmark & $\times$  \\ \hline
        \cite{safaei2019vlcc} & \checkmark & $\times$  & $\times$  &  \checkmark &  $\times$ \\ \hline
        \cite{yuan_prediction_2021} & $\times$  & $\times$  & $\times$  & \checkmark & \checkmark \\ \hline
    \end{tabular}
  
    \label{tab:data_fusion_papers}
\end{table}

\subsubsection{Data Fusion for FOC}
The data related to FOC come from different sources. These sources may share common variables, meaning one feature type is available across different data sources. Each source has its unique advantages and limitations. For example, the data coming from onboard sensors has high resolution. In addition, a feature recorded in a data source may be more accurate than another. For instance, the observations recorded on onboard sensors for wind magnitude are more accurate than the open-source meteorological data available. The works on data fusion for the estimation of FOC include
\citet{li_data_2022, du_data_2022, wang2023ship, ren2022container, hajli_fuel_2024, zhang2023research, zhu2021modeling}.
A comparison of different works based on the fusion of different data sources is presented in Table \ref{tab:data_fusion_papers}. The most common approach is to fuse the NRs with metocean data for FOC prediction, as long voyages last several days.  When real-time fuel consumption prediction is required, data from onboard sensors is fused with data from other sources. There are several takeaways from the data fusion for FOC estimation purposes. A common way for FOC is to fuse noon reports with metocean data from publicly available sources.

\subsubsection{Performance Metrics}

According to the literature, the performance of data-driven models for FOC estimation is typically evaluated using standard regression metrics, including:
Mean Absolute Error (MAE) ( $\frac{1}{n} \sum_{i=1}^n |y_i - \hat{y}_i|$), Mean Square Error (MSE) ($\frac{1}{n} \sum_{i=1}^n (y_i - \hat{y}_i)^2$) Root Mean Square Error (RMSE) ($\sqrt{\frac{1}{n} \sum_{i=1}^n (y_i - \hat{y}_i)^2}$), Mean Absolute Percentage Error (MAPE), ($\frac{100}{n} \sum_{i=1}^n \left |\frac{y_i - \hat{y}_i}{y_i} \right|$) and the Coefficient of Determination ($R^2$) ($1 - \frac{\sum_{i=1}^n (y_i - \hat{y}_i)^2}{\sum_{i=1}^n (y_i - \bar{y})^2}$),  where $\bar y$ is the sample mean of observations, given by $\bar y = \frac{1}{n} \sum_{i=1}^n y_i$.. 
These metrics measure the deviation between observed and predicted fuel consumption values and are widely used across the reviewed studies \cite{fan_comprehensive_2024, du_two-phase_2019, yuksel_comparative_2023, yan2024improving}

\begin{table}[H] % Forces table to stay exactly here, on a new page
    \centering
     \caption{Summary of the data-driven papers for FOC estimation in maritime.}
    \renewcommand{\arraystretch}{1.5}
    \begin{tabular}{|p{1.8cm}|p{2.8cm}|p{5.8cm}|p{2.0cm}|p{2.0cm}|}
        \rowcolor{gray!20}
        \hline
        \textbf{Paper} & \textbf{Output} & \textbf{Features Included} & \textbf{Method Type} & \textbf{Results} \\ \hline
        \cite{hajli_fuel_2024} & Metric ton per nautical mile & rpm, sog, slip, wind factor, averaged wind factor, wave factor, averaged wave factor, current factor, averaged current factor & multiple linear regression & MAE: $5.49 \times 10^{-3}$, RMSE: $7.36\times 10^{-3}$ (metric tons per nautical mile) \\ \hline
        \cite{du_two-phase_2019} & Metric ton per day & Speed, displacement, wave height, wave direction, wind force, current speed, current direction, sea water temperature, trim & ANN & RMSE of less than 9.5 MT/day \\ \hline
        \cite{papandreou2022predicting} & FOC (main engine) [tn] & Speed through water, Relative wind direction, Relative wind speed, Mean draft, Trim, Days since last drydock, Laden or Ballast (binary format) & XGBoost & prediction error with 5\% \\ \hline
        \cite{yan2024improving} & Hourly average ME fuel consumption rate & average sailing speed, sailing condition (laden/ballast), wind direction, wind force, draft, air density over the oceans, forecast surface roughness, mean sea level pressure, mean direction of total wave, mean period of total wave, significant height of the total wave, temperature of water near sea surface & ANN & MAE of 7.5\% \\ \hline
        \cite{uyanik_machine_2020} & FOC (ton/day) & M/E fuel oil inlet pressure (bar), temperature (°C), viscosity (cSt), Air coolers cooling water temperature (°C), Jacket cooling water inlet pressure (bar) and temperature (°C), Scavenging air receiver temperature (°C), Bearing temperatures (°C), Start air inlet pressure (bar), Fuel mass flow (kg/h), Shaft power (kW), rpm, torque (kNm), M/E lubrication oil temperature (°C) and pressure (bar), M/E fuel flow counter, M/E power, Exhaust temperature of the cylinders (°C), Turbocharger rpm and exhaust gas inlet temperature (°C). & Bayesian ridge regression, Kernel ridge regression, Multiple linear regression and Ridge regression & Bayesian Ridge (RMSE=0.0001 MAE=0.003 $\text{R}^2$=99.999\%), Multiple LR \& Ridge (RMSE=0.0001 MAE=0.002 $\text{R}^2$=99.999\%), Kernel Ridge (RMSE=0.0001 MAE=0.003 $\text{R}^2$=99.999\%) \\ \hline
        \cite{fan_comprehensive_2024} & ME FOC (kg/h) & Wind speed/direction, water depth, speed (over ground/water), engine fuel use/power, propeller speed/power, bearing temp, exhaust temp/pressure, intercooler air temp, cylinder liner water temp, fuel pressure/temp, and exhaust temp after vortex. & Random Forest and Extreme Gradient Boosting & RF ($\text{R}^2$=0.9747, MAE=1.72) \\ \hline
        \cite{yuksel_comparative_2023} & ME FC (tonnes) & wind direction, wind force, sea state, swell height, distance, speed, slip & M5--decision tree with linear regression leaves for continuous forecasting.  & M5 ($\text{R}^2$=0.9666) \\ \hline
    \end{tabular}
   
    \label{tab:datadriven-papers}
\end{table}

\clearpage

\begin{table}[h]
    \centering
    \renewcommand{\arraystretch}{1.5}
    \begin{tabular}{|p{3.7cm}|c|c|c|c|c|}
        \rowcolor{gray!20}
        \hline
        \textbf{Papers} & \textbf{Engine Data} & \textbf{Weather/Ocean Data} & \textbf{Operational Data} & \textbf{ML Method} & \textbf{Tree-based} \\
        \hline
        \cite{hajli_fuel_2024} & $\times$ & \checkmark & \checkmark & $\checkmark$ & $\times$ \\ \hline
        \cite{du_two-phase_2019} & $\times$ & \checkmark & \checkmark & \checkmark & $\times$ \\ \hline
        \cite{papandreou2022predicting} & $\times$ & \checkmark & \checkmark & \checkmark & \checkmark \\ \hline
        \cite{yan2024improving} & $\times$ & \checkmark & \checkmark & \checkmark & $\times$ \\ \hline
        \cite{uyanik_machine_2020} & \checkmark & $\times$ & \checkmark & \checkmark & \checkmark \\ \hline
        \cite{fan_comprehensive_2024} & \checkmark & \checkmark & \checkmark & \checkmark & \checkmark \\ \hline
        \cite{yuksel_comparative_2023} & $\times$ & \checkmark & \checkmark & \checkmark & \checkmark \\ \hline
    \end{tabular}
    \caption{Summary of data-driven papers for FOC estimation with their data types and methods. "Engine Data" refers to onboard sensor readings; "Operational Data" includes voyage characteristics such as speed, trim, or draft; "Weather/Ocean Data" refers to environmental inputs.}
    \label{tab:foc_data_sources_methods}
\end{table}

\subsection{Hybrid Models for Fuel Consumption Estimation} 
{Hybrid or Grey-Box Models (GBMs) combine physics-based (white-box) and data-driven (black-box) approaches to leverage the strengths of both paradigms \cite{rai2020driven}, \cite{thuerey_physics-based_2022}, \cite{zhang_physics-infused_2021}.
Examples of GBMs for the estimation of FOC in the maritime domain include \cite{odendaal2023enhancing,dekeyser2022towards,meng_shipping_2016,yang2019genetic, coraddu_vessels_2017, du_two-phase_2019, tu_optimum_2023, kaneko_hybrid_2023, agand_fuel_2023, liu2020voyage}.

To ensure consistency and clarity, we adopt the following standardized terminology for GBMs based on how the physics-based and machine learning components interact:

\begin{itemize}
    \item \textbf{Serial Grey-Box Models (S-GBM):} The physics-based model is executed first, and its output (e.g., estimated resistance or power) is used as input to a machine learning model. The ML component refines residual errors or nonlinearities that the physics model cannot capture. 
    
    \item \textbf{Constraint-based Grey-Box Models (C-GBM):} The physics knowledge is incorporated as mathematical constraints or regularization terms during ML training. This ensures that predictions remain physically consistent while allowing the data-driven model to learn flexible relationships within defined boundaries.
    
    \item \textbf{Embedding-based Grey-Box Models (E-GBM):} The physics relationships or equations are embedded directly into the ML architecture itself, often replacing a computationally expensive physics solver. This approach enables real-time estimation but may reduce interpretability.
\end{itemize}

These distinctions remove the conceptual overlap in earlier drafts and align the terminology with recent hybrid modeling frameworks \cite{coraddu_vessels_2017,dekeyser2022towards,guo_combined_2022}. Based on how the physics-based and machine-learning components interact as described in Table \ref{tab:gbm_comparison}.

\subsubsection{Serial Grey-Box Models (S-GBM)}
In this type of GBM, the output of the physics-based models serves as input to the data-driven approach, such as a neural network. This is the most common type of GBM for the estimation of FOC. This type of model is also referred to as a combination approach of GBM \cite{ruan2024novel,liu2020voyage}. That is because it requires the sequential integration of physics-based and ML models. For example, in \citet{dekeyser2022towards}, various additional resistances are calculated, and a neural network is employed to determine the relationship between the power and speed of the vessel.
\citet{zeng2022data} combines physics-based empirical models with machine-learning models to make an ensemble model of machine learning for the FOC of ships. In \cite{ma2023interpretable}, a GBM approach for the prediction of FOC is presented, which improves the interpretability of the model, quantitatively demonstrates the influence of input features on the fuel consumption of the vessel, and further validates the effectiveness of WBM on improving the performance of the GBM prediction.

\subsubsection{Constraint-Based Grey-Box Models (C-GBM)}
This approach is sometimes referred to as the constraint approach. It utilizes physics-informed regularization. In this GBM, some a priori information is included in the data-driven model. For instance, the regularization of the data-driven model is modified to incorporate domain knowledge. In \cite{coraddu_vessels_2017}, an approach called advanced grey-box model (A-GBM), where a ridge regularization of the difference between the GBM parameters and WBM parameters is included to accommodate a priori information about the problem. 

\subsubsection{Embedding-Based Grey-Box Models (E-GBM)}
Embedding approaches involve hybrid computational architectures, where some computationally expensive part of the physics-based model is replaced by a data-driven model. Although such hybrid models exist in other applications, they are rarely used for fuel consumption in maritime settings. Technically, \citet{guo_combined_2022} can be considered a hybrid model, as machine learning is employed in some parts of the model.

\begin{table}[ht]
    \centering
    \caption{Comparison of Grey-Box Modeling (GBM) approaches across physics integration, machine learning role, advantages, and weaknesses.\\
    \textbf{[1]}~\cite{dekeyser2022towards},
    \textbf{[2]}~\cite{coraddu_vessels_2017},
    \textbf{[3]}~\cite{guo_combined_2022}}
    \renewcommand{\arraystretch}{2.8} % For better row spacing
    \rowcolors{2}{gray!10}{white}
    \begin{tabularx}{\textwidth}{ p{0.6cm} p{1.7cm} p{3.0cm} p{2.4cm} p{2.5cm} p{4.0cm} }
        \toprule
        \rowcolor{gray!30}
        \textbf{Paper} & \textbf{GBM Type} & \textbf{Physics-Based Role} & \textbf{ML Role} & \textbf{Key Advantage} & \textbf{Weaknesses} \\
        \midrule
        \textbf{[1]} & \textbf{Serial (S-GBM)} & Input for ML model & Corrects residual errors & Balances accuracy \& interpretability & Performance depends on quality of the physics-based model; limited adaptability if underlying physics is inaccurate. \\
        
        \textbf{[2]} & \textbf{Constraint (C-GBM)} & Encodes physical laws as constraints during ML training & Learns within physics-informed boundaries & Ensures physically valid outputs & Requires careful formulation of constraints; may restrict flexibility and increase training complexity. \\
        
        \textbf{[3]} & \textbf{Embedding (E-GBM)} & Embeds physics components directly in ML architecture & Replaces computationally expensive physics solvers & Enables fast, real-time estimation & Implementation is complex; may lose interpretability and risk overfitting without sufficient physical guidance. \\
        \bottomrule
    \end{tabularx}
    \label{tab:gbm_comparison}
\end{table}

\begin{comment}

\begin{table}[ht]
    \centering
    \renewcommand{\arraystretch}{2.5}
    \rowcolors{2}{gray!10}{white}
    \begin{tabularx}{\textwidth}{l X X  p{2.5cm} p{1.5cm}}
        \toprule
        \rowcolor{gray!30}
        \textbf{GBM Type} & \textbf{Physics-Based Role} & \textbf{ML Role} & \textbf{Key Advantage} & \textbf{Paper} \\
        \midrule
        \textbf{Serial (S-GBM)} & Input for ML model & Corrects residual errors & Balances accuracy \& interpretability & \cite{dekeyser2022towards} \\
        \textbf{Constraint (C-GBM)} & Constraints in ML training & Learns within physics limits & Ensures physically valid outputs &  \cite{coraddu_vessels_2017} \\
        \textbf{Embedding (E-GBM)} & Embedded in ML architecture & Replaces physics solvers & Enables real-time estimation & \cite{guo_combined_2022} \\
        \bottomrule
    \end{tabularx}
    \caption{\st{Comparison of Grey-Box Modeling Approaches.} \revision{Comparison of Grey-Box Modeling (GBM) approaches across physics integration, machine learning role, and advantages.}}
    \label{tab:gbm_comparison}
\end{table}
    
\end{comment}

\subsection{Explainable AI (XAI) for FOC Estimation}
Machine learning (ML) models have demonstrated high accuracy in predicting FOC. However, most black-box models, such as deep learning, suffer from poor interpretability, making them difficult to trust for decision-making in maritime operations. Many studies have incorporated XAI techniques to address this issue to improve transparency while maintaining predictive accuracy \cite{lundberg2017unified}.

With the growing adoption of AI-driven models for estimating ship fuel consumption, the lack of transparency in black-box models has become a significant challenge \cite{ma2023interpretable}. Maritime stakeholders, including ship operators and regulatory bodies, require interpretable models to ensure compliance with fuel efficiency regulations and environmental sustainability goals. XAI techniques provide a means to improve model transparency, making AI-based predictions more understandable and trustworthy, as shown in Figure \ref{fig:xai_methods_in_foc_estimations}.

Furthermore, integrating XAI into regulatory compliance frameworks can enhance the acceptance of AI-based fuel optimization solutions. By clearly explaining model predictions, ship operators can make informed decisions on route planning, speed adjustments, and fuel efficiency strategies. Future research should explore the development of hybrid models that incorporate both data-driven AI techniques and traditional maritime engineering principles while ensuring transparency and interpretability through XAI.

Several studies have explored different XAI methods, each showcasing a different level of interpretability and associated trade-offs with accuracy as illusted in Table \ref{tab:XAI_Comparison}, \cite{ma2023interpretable}, \cite{wang2023innovative}, \cite{handayani2023navigating}, \cite{handayani2025predictive}:

\begin{table}[h]
    \centering
    \caption{Comparison of XAI Approaches in FOC Estimation.\\
     \textbf{[1]} \cite{ma2023interpretable},
    \textbf{[2]} \cite{wang2023innovative},
    \textbf{[3]} \cite{handayani2023navigating},
    \textbf{[4]} \cite{handayani2025predictive}
    }
    \renewcommand{\arraystretch}{1.3} % Adjust row height
    \begin{tabular}{|p{0.7cm}|l|p{4.0cm}|p{4.0cm}|p{3.0cm}|}
      \rowcolor{gray!20}
        \hline
        \textbf{Paper} & \textbf{XAI Method} & \textbf{Interpretability} & \textbf{Accuracy} & \textbf{Key Trade-Off} \\ 
        \hline
        \textbf{[1]} & SHAP & High  & Moderate  & Loss of some model flexibility    \\ 
        \hline
        \textbf{[2]} & PI-NN \& MIO & PI-NN: Moderate, MIO: High  & PI-NN: High, MIO: Moderate    & MIO is highly interpretable but less accurate    \\ 
        \hline
        \textbf{[3]} & SHAP + XGBoost & High  & High (R² = 0.95)  & Computational overhead due to SHAP analysis    \\ 
        \hline
        \textbf{[4]} & SHAP + XGBoost & High  & Very High (R² = 0.99)  & Model sensitivity to different load conditions    \\ 
        \hline
    \end{tabular}
    
    \label{tab:XAI_Comparison}
\end{table}

\begin{comment}
\begin{table}[h]
    \centering
    \renewcommand{\arraystretch}{1.3} % Adjust row height
    \begin{tabular}{|p{2.0cm}|l|p{2.5cm}|p{4.0cm}|p{3.0cm}|}
      \rowcolor{gray!20}
        \hline
        \textbf{Paper} & \textbf{XAI Method} & \textbf{Interpretability} & \textbf{Accuracy} & \textbf{Key Trade-Off} \\ 
        \hline
        \textbf{\cite{ma2023interpretable}} & SHAP & High  & Moderate  & Loss of some model flexibility    \\ 
        \hline
        \textbf{\cite{wang2023innovative}} & PI-NN \& MIO & PI-NN: Moderate, MIO: High  & PI-NN: High, MIO: Moderate    & MIO is highly interpretable but less accurate    \\ 
        \hline
        \textbf{\cite{handayani2023navigating})} & SHAP + XGBoost & High  & High (R² = 0.95)  & Computational overhead due to SHAP analysis    \\ 
        \hline
        \textbf{\cite{handayani2025predictive}} & SHAP + XGBoost & High  & Very High (R² = 0.99)  & Model sensitivity to different load conditions    \\ 
        \hline
    \end{tabular}
    \caption{\revision{Comparison of XAI Approaches in FOC Estimation.}}
    \label{tab:XAI_Comparison}
\end{table}

\end{comment}

\begin{figure}[!ht]
    \centering
    \includegraphics[scale=0.12]{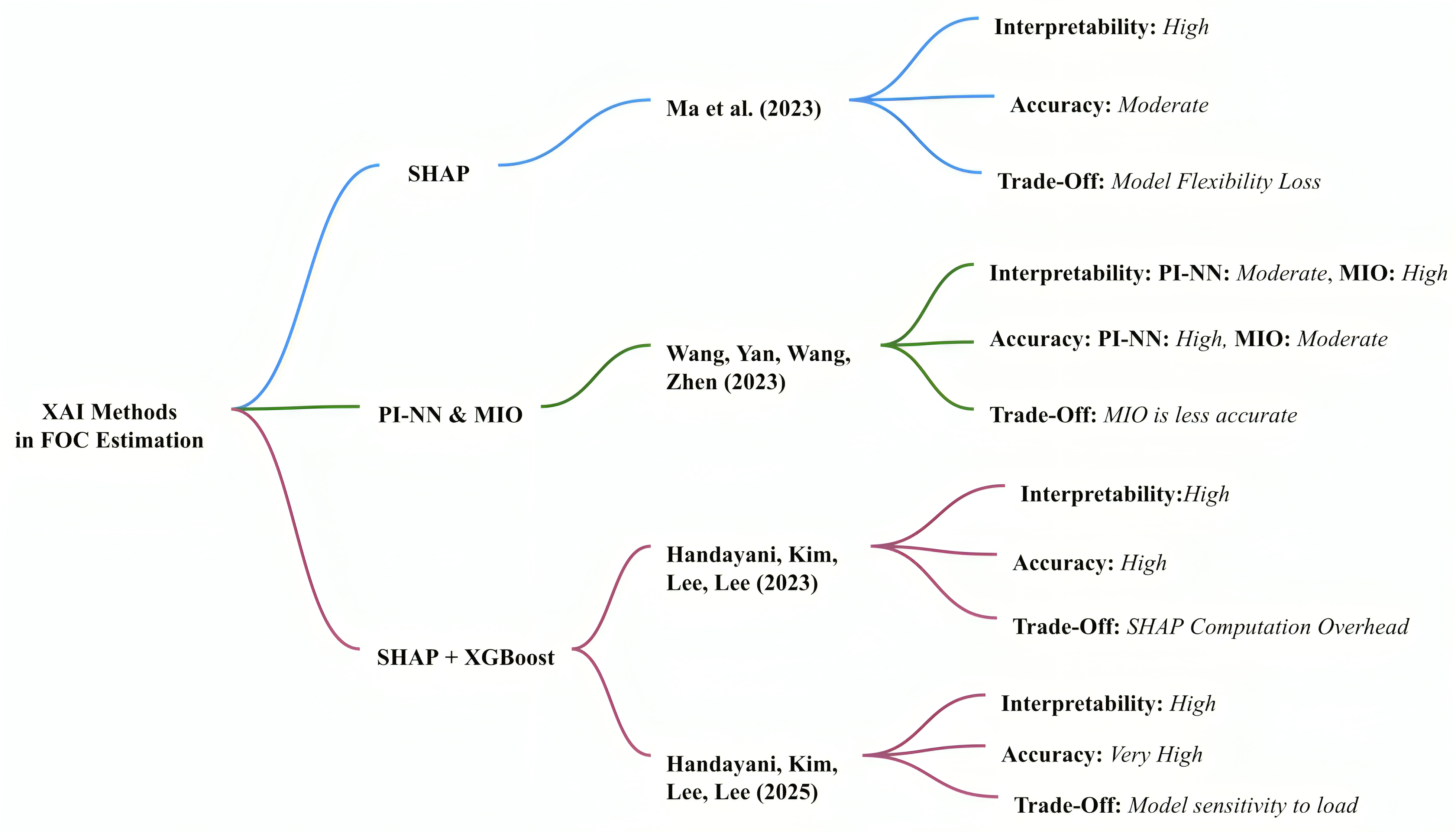}
    \caption{XAI Methods in FOC Estimation}
    \label{fig:xai_methods_in_foc_estimations}
\end{figure}

\subsubsection{SHAP (SHapley Additive exPlanations) for Feature Importance }

SHAP is a prominent method for explaining machine learning model predictions by attributing importance scores to input features. In FOC estimation, SHAP was utilized with a Random Forest model to evaluate the influence of operational and environmental factors, like speed and weather, on fuel consumption \cite{wang2023innovative}. The RF model was developed using a dataset of ship performance metrics, encompassing variables such as speed, trim, hull fouling, and weather data. Subsequently, SHAP values were calculated to determine the marginal impact of each feature on FOC predictions, thereby identifying significant factors across various ship voyages. A sensitivity analysis was performed by modifying key operational parameters to observe changes in SHAP values \cite{ma2023interpretable}.

The results showed that speed and trim adjustments were the most significant factors affecting FOC, followed by weather conditions such as wind speed, wave height, and ocean currents, which played a crucial role, especially in long-haul voyages. Additionally, ship hull condition, particularly biofouling levels, was another critical variable influencing deviations in FOC since increased hull resistance led to higher propulsion power demands. By using SHAP, ship operators can interpret how changes in operational settings affect fuel consumption, providing a transparent and interpretable decision-support framework for optimizing fuel efficiency. This enabled them to proactively adjust vessel parameters based on real-time environmental conditions, enhancing operational decision-making \cite{ma2023interpretable}.

Despite its strengths, using SHAP-based models came with trade-offs. While SHAP provided high interpretability, the RF model's predictive accuracy was slightly lower compared to deep learning approaches such as LSTMs or transformers, which better capture highly non-linear relationships in the data. However, the computational efficiency of SHAP-based models made them more suitable for real-time maritime applications, unlike deep learning models, which often require extensive training and computational resources. Thus, while SHAP is an effective tool for improving the interpretability of the model in FOC estimation, a balance exists between interpretability and predictive performance \cite{ma2023interpretable}.

\subsubsection{Physics-Informed Neural Networks (PI-NN) vs. Mixed-Integer Optimization (MIO) }

Physics-Informed Neural Networks (PI-NN) and Mixed-Integer Optimization (MIO) represent two different approaches to balancing accuracy and interpretability in FOC estimation and optimization. PI-NNs incorporate physical constraints, such as ship resistance equations, propulsion efficiency, and fuel-power-speed relationships, into machine learning models \cite{wang2023innovative}. By embedding these constraints, the neural network ensures that predictions remain physically plausible, reducing the likelihood of non-physical outputs. On the other hand, MIO takes a structured optimization approach, formulating fuel consumption minimization as a constrained optimization problem using linear and quadratic constraints. Unlike machine learning models, MIO does not rely on large training datasets but instead uses predefined constraints to solve an optimization problem at each decision point \cite{handayani2023navigating}  \cite{wang2023innovative}.

Moreover, findings demonstrated that PI-NNs offered a balance between accuracy and interpretability, making them highly effective in real-world maritime applications. Since PI-NNs enforce physical consistency, their predictions align more closely with established maritime principles, enabling improved generalization across various ship types and operational conditions. In contrast, MIO provided full transparency, ensuring that all fuel optimization decisions were fully interpretable. However, its predictive power was lower than that of PI-NNs, particularly in complex, high-dimensional scenarios where multiple interacting variables influence fuel consumption. PI-NNs performed better in dynamic maritime environments, where fuel efficiency is influenced by rapidly changing sea states, wind patterns, and operational adjustments, whereas MIO was more rigid in handling such uncertainties \cite{handayani2023navigating}.

\subsubsection{LIME: Interpretable Fuel Oil Consumption Prediction in Cargo Container Vessels  }
The study by \cite{handayani2023navigating} focuses on enhancing the interpretability of FOC prediction in cargo container vessels by integrating machine learning models with Explainable Artificial Intelligence (XAI) techniques. They developed an XGBoost Regressor model to predict FOC in cargo container vessels, achieving high accuracy as shown in Table \ref{tab:XAI_Comparison} and Figure \ref{fig:xai_methods_in_foc_estimations}. The study applied SHAP to enhance interpretability, identifying ‘Average Draught (Aft and Fore)’ as the most significant operational factor and ‘Relative Wind Speed’ as the dominant environmental influence.

The study also examined regional FOC variations, highlighting extreme fuel consumption in the Strait of Malacca and the South China Sea, demonstrating the impact of geographic conditions. By integrating XAI techniques, the research provides actionable insights to optimize ship operations, enhance energy efficiency, and ensure regulatory compliance in maritime transport.

\subsubsection{Predictive analysis of fuel oil consumption in vessels}
\cite{handayani2025predictive} developed a predictive model for FOC in vessels, emphasizing the role of load conditions (laden, ballast, and empty) using XGBoost Regressor. The model achieved an $R^2$ of 0.99, optimizing FOC prediction across various operational and environmental conditions.

The study employed XAI techniques, such as SHAP, to investigate how both operational (speed, draft, and trim) and environmental (wind, wave, and sea conditions) factors influence FOC. It found that the average draft was the most influential operational factor, while relative wind angle significantly impacted fuel consumption under laden conditions.

The research highlights the importance of accounting for different loading conditions when predicting FOC, as the influence of operational and environmental factors varied significantly across these states. By applying XAI, the study provides actionable insights to improve fuel efficiency and reduce emissions, thereby contributing to sustainable shipping practices.

\subsubsection{Limitations of XAI in Maritime Contexts}
While XAI methods have shown great promise in improving the interpretability of data-driven and hybrid models for FOC estimation, several limitations hinder their widespread adoption in maritime applications. Firstly, most XAI techniques, such as SHAP and LIME, are \textit{post-hoc} interpretability methods, meaning they explain model predictions after training rather than ensuring transparency during model construction. Consequently, explanations may not always reflect the model's true internal reasoning, which can mislead users in operational contexts \cite{ma2023interpretable, wang2023innovative}. Secondly, the maritime domain suffers from limited labeled and high-quality data, making it difficult to generate reliable feature importance measures or explanations across diverse vessel types and environmental conditions \cite{handayani2023navigating}. The lack of standardized datasets and benchmarking protocols further limits the reproducibility and comparability of XAI-based findings. 

Furthermore, there exists a fundamental trade-off between \textit{model accuracy and interpretability}. Highly interpretable models (e.g., decision trees, mixed-integer optimization) often perform worse on complex nonlinear relationships, while deep learning models that achieve higher accuracy are less transparent \cite{handayani2025predictive, wang2023innovative}. This tension is particularly critical in safety-sensitive maritime operations, where trust and explainability are prerequisites for regulatory acceptance. Finally, integrating XAI outputs into decision-support systems remains challenging. While XAI can highlight influential features, it does not directly translate these insights into actionable recommendations for navigation or engine control. Bridging this gap requires interdisciplinary approaches that combine domain expertise, human-in-the-loop feedback, and regulatory frameworks to ensure interpretable, reliable, and auditable maritime AI systems \cite{ma2023interpretable, wang2023innovative}.

\section{Fuel Oil Consumption Optimization}
As discussed in the article's introduction, it is crucial to reduce FOC for both environmental and financial reasons. There are certain ways to improve the FOC of ships, including design and operational measures \cite{elkafas2022advanced}. 

First, we discuss a categorization of the nature of the FOC optimization problem. Later, we discuss the classification of approaches to solve the problem of FOC optimization
\cite{karakostas_enhanced_2024}.

\subsection{Nature of the Problem of FOC Minimization}
The problem of optimization of FOC can be categorized into the following main categories based on the nature of the problem. 

\subsubsection{Vessel-parameter Optimization}
In this category, FOC optimization approaches are proposed to optimize certain parameters of ships, such as speed, trim, rpm, and draft, for a given route.
The most common way to optimize fuel consumption is to maximize the speed of the vessel for a given route, as investigated in \cite{yang2020ship, medina2020bunker, gao2021speed, zheng_voyage_2019, yan2020development}. According to \citet{taskar2023case}, speed optimization for ships can save up to 6\%. Additionally, speed optimization can help avoid adverse weather conditions at sea and achieve the required ETA. Trim of the ship is also considered in the literature, and trim optimization can save fuel by 2--3\%, according to \cite{reichel2014trim}. The work in \cite{islam2019effect} demonstrates how the resistance prediction varies for the ship model under different trim conditions, at varying speeds and drafts of the ship. Thus, adjusting the trim would affect the ship's FOC.
\citet{hu_two-step_2022} presents a solution to optimize the trim of the ships. Keeping the features constant, the optimal value of the trim is computed by dividing the possible values using regular intervals and calculating the FOC for each trim value. Other works investigating trim optimization include \cite{elkafas2022advanced}.

Each of these optimization categories can be associated with representative algorithms commonly adopted in the literature:

\begin{itemize}
    \item \textbf{Vessel-parameter optimization:} typically solved using analytical or heuristic algorithms such as Genetic Algorithms (GA) \cite{khan_benefits_2022, ma2023interpretable}, Particle Swarm Optimization (PSO) \cite{zheng_voyage_2019}, or regression-based gradient search methods \cite{hu_two-step_2022}. These algorithms adjust controllable parameters (e.g., speed, trim, RPM) to minimize FOC under given voyage constraints.
    
    \item \textbf{Environment-based optimization (weather routing):} often handled using Dynamic Programming (DP), Multi-Objective Evolutionary Algorithms (MOEA), or Reinforcement Learning (RL) frameworks \cite{moradi_marine_2022, szlapczynska_preference-based_2019, du_data_2022}. These approaches find optimal routes or speed profiles while considering environmental and weather conditions.
    
    \item \textbf{Holistic optimization:} combines both vessel and environmental parameters, typically using hybrid methods such as Reduced Space Search Algorithms (RSSA), multi-level GA–PSO hybrids, or heuristic–data-driven frameworks \cite{ormevik2023high, yuan_prediction_2021, sang_ship_2023} . These approaches jointly optimize routing, propulsion, and voyage scheduling.
\end{itemize}

Several works consider optimization of FOC w.r.t. speed and trim jointly, e.g., \cite{du_two-phase_2019, fan2022joint, xie2023joint, wang2023novel}. The vessel parameter optimization problem can be solved using single objective-driven voyage optimization techniques \cite{yu_literature_2021}. For example, \cite{luo2023comparison} investigates the effects of weather parameter forecasts when optimizing the overall fuel consumption of the voyage w.r.t. the speed of the ship. 
\subsubsection{Environment-based Optimization/Weather Routing}
Weather routing has become very popular recently in the maritime industry \cite{zis2020ship}, 
\cite{gkerekos_novel_2020},
\cite{du_energy_2022}.
The concept of weather routing involves calculating an optimal route for a given pair of sources and destinations, based on specific criteria that take into account environmental conditions (e.g., water depth) and weather conditions (including wind, waves, and sea currents).
Note that in such optimization methods, the estimation of FOC utilizes either a data-driven model or a physics-based model. For example, in \cite{du2021estimation}, the physics-based FOC estimation model is used to calculate the optimal route with minimum fuel consumption under constraints such as ETA, the power rating, and safety of the main engine, and the maximum speed limit. \cite{szlapczynska_preference-based_2019} presents a solution based on evolutionary multi-objective optimization (EMO) for weather routing. The goal is to find optimal routes by optimizing passage time, FOC, and safety by considering weather, ship characteristics, and safety threats to the ship.
%\subsubsection{Trim Optimization}
%\subsubsection{Speed/rpm Optimization}

\subsubsection{Holistic Optimization}
This category considers both the optimization of vessel parameters and the environment-based optimization. This means that optimal routes and the optimal parameters for vessels are also computed. This is rather challenging; however, there are works available in the literature that fall into this category \cite{vettor2015multi}. In \cite{ormevik2023high}, a scheduling problem involving fuel consumption is addressed on a given route, where speed optimization is considered under various weather conditions using different approaches to estimate the Fuel Optimal Control and investigate its influence on speed optimization strategies. Data-driven models are used to estimate FOC taking into account environmental factors, and the engine speed is optimized using the reduced space search algorithm (RSSA) in \citet{yuan_prediction_2021}. \cite{du_energy_2022} also optimizes FOC and ETA by finding an optimal route given the weather conditions and constraints. \cite{sang_ship_2023} also addresses the problem of optimizing FOC under different operating conditions and voyage environments and provides suggestions about the propulsion, maneuver control, voyage planning, and engine management of the ship to reduce fuel consumption.  
\cite{ma_method_2020} jointly optimizes the route and speed of the ship to minimize FOC and ultimately reduce emissions within emission control areas (ECAs). A similar type of problem is addressed in \cite{zhao_bi-objective_2019} with different types of fuel to comply with the regulations of the ECAs. \cite{veneti2017minimizing} addresses and investigates the role of weather routing in reducing both fuel consumption and maritime risk.

\subsection{Approaches as a Solution to FOC Minimization}
The solutions of the FOC minimization problem can be divided into the following categories. Real-time optimization of the energy efficiency of the ship is considered in \cite{wang2016real}, where the optimal engine speed is calculated for a given set of environmental conditions. A similar problem is solved in \cite{yan2018energy}, where the optimal engine speed is computed for each section of the route given the environmental factors and conditions. In \cite{moradi_marine_2022}, a reinforcement-based solution is proposed for route optimization to find a route with the minimum FOC and the corresponding speed and course for given environmental conditions. 

\subsubsection{Analytical/Heuristic Approaches}

In this category of approaches for FOC optimization, the formulated optimization problem is solved using analytical or heuristic methods. 

\cite{khan_benefits_2022} provides a benchmark for the use of the evolutionary genetic algorithm for voyage optimization (both fuel consumption and ETA). A three-objective minimization problem is considered, that is, fuel consumption, voyage time, and voyage distance.  The methodology involves the following three steps:
i) Route representation: The possible sailing area between the source and destination is divided into a mesh of 250 nodes that act as waypoints. The speed parameter is maintained between two waypoints. The speed is maintained at a constant rate of four waypoints. 
ii) Fuel consumption: The fuel is predicted according to ISO15016:2015. This model calculates the resistance of the ship in calm water and the additional resistance due to weather conditions, including waves, winds, and currents. The total resistance of the ship can be used to calculate the effective power required, which in turn determines the fuel consumption of the ship. 
iii) Safety and voyage constraints: To ensure safety, the land areas are removed from the search space. Maximum values for wind speed, wave height, turning angle, and engine power have been established. The traffic separation scheme is also included in the problem formulation.
Using a similar approach, another important factor is added to the analysis in \cite{ma2023multi}, which is the ship's sailing state. The idea is that the ships' sailing state provides important information about the sea conditions. The sea conditions are clustered, and for each segment of the route, the main engine speed is optimized through PSO, considering both the ship's fuel consumption and sailing time.

In \cite{gkerekos_novel_2020}, a data-driven heuristic framework for weather routing is proposed, which takes into account the historical performance of the ship and the current weather conditions on a discretized grid of points and includes a data-driven FOC prediction model based on neural networks together with a modified Dijkstra algorithm to find the optimal route. The PSO based algorithm is proposed in \cite{zheng_voyage_2019} under the assumptions that no bad weather will occur and the arrival time range and docking time are predetermined, given i) the number of stations and their coordinates, ii) the arrival time range and the docking time of each station, iii) the sailing period of a voyage, and iv) the change of speed and load during the previous voyage. The goal is to determine i) the distribution of speed and load, the expectation of sailing speed between every two stations, and the total fuel consumption. ANNs are used to estimate the fuel consumption of the ship. A multi-objective evolutionary algorithm based on decomposition (MOEA/D) is proposed in \cite{zakerdoost2019multi} for FOC optimization. Other examples of genetic algorithms or PSO-related works for FOC optimization include \cite{wang_voyage_2021}, \cite{kuhlemann_genetic_2020}, and \cite{yan2018energy}.

\subsubsection{Data-oriented Approaches}
These methods utilize data from various sources and employ a data-centric approach to estimate FOC within the FOC optimization solution. These solutions are based on different techniques for FOC modeling. For example, \cite{yuan_prediction_2021} exploits LSTMs for fuel consumption prediction inside the FOC optimization solution. ANNs are used inside FOC optimization to predict fuel consumption in \cite{moradi_marine_2022}. Similarly, \cite{rudzki2022optimization} utilizes ANNs for estimation within the two-criteria optimization model, which involves minimizing both ship fuel consumption and navigation time.

\subsubsection{Mixed Approaches}
These methods involve combining data-centric and WBMs to estimate FOC within the FOC optimization solution. An example of such FOC optimization models is \cite{zwart2023grey}, which utilizes a GBM. Data from noon reports are used. Another similar work is \cite{yang_genetic_2019}, which utilizes a GBM and obtains data from onboard sensors. Trim and main engine power are jointly optimized in \cite{yu2024trim} while keeping the voyage length and time fixed. 

\section{Discussions}
Due to the complex nature of the problem, estimating and optimizing FOC for maritime vessels remains an open challenge. In this paper, we discuss the challenges of estimating and optimizing FOC. Moreover, the limitations of existing work on FOC estimation and optimization are also discussed. 

\subsection{Challenges}
Estimation and optimization of fuel consumption in maritime transport is a highly complex task, as several factors affect FOC of a ship in one way or another \cite{merien2018situ}. At the same time, the available works on fuel consumption estimation and optimization are pretty effective, and several cases have been recorded in which they have been implemented and deployed on ships. However, the available work has challenges and limitations.  We distinguish between technical challenges, which relate to data quality, modeling, and algorithmic limitations, and operational challenges, which concern real-time deployment, regulatory constraints, and practical adoption in vessels. The following points describe both categories, highlighting where each type of challenge arises and how they interact in real-world maritime contexts.

\subsubsection{Data-related Challenges}
A major bottleneck in maritime fuel analysis is data accessibility, quality, and heterogeneity. While AIS and meteorological data are publicly available, high-resolution onboard sensor and MRV datasets remain proprietary, limiting research reproducibility \cite{luo2023after, heikkila2024effect}. The available data are often noisy, incomplete, or inconsistent across ship types, which complicates model calibration and validation \cite{merien2018situ}. Furthermore, the lack of standardized and open benchmark datasets hinders fair comparison of algorithms and slows progress toward generalizable solutions \cite{zhu2021modeling, li_data_2022, guo_combined_2022}.

\begin{itemize}
    \item Data availability: While AIS data is publicly available, limited public access to FOC-related data restricts research and model development, as proprietary datasets are often controlled by shipping companies. Therefore, data related to the fuel consumption of ships in maritime is typically not available for research purposes.
    \item Data quality: The data in maritime can be noisy and can contain missing, erroneous, or invalid values. This also applies to FOC-related data.  For example, noon reports are manually entered by the ship crew; some values can be missing or invalid.
    \item Data quantity: Some data-driven approaches need a lot of data. However, obtaining data on the maritime industry can be a challenge, as the data belongs to the ship's owners, and they may be reluctant to share it.
    \item Data diversity: There are multiple types of vessels used for different purposes. These vessels have very different features and systems. Therefore,  the available data related to FOC may only belong to one type of ship. Hence, data are not always available for training and testing data-driven approaches for all other types of ships.
 \end{itemize}

\subsubsection{Modeling and Explainability}
Technical challenges relate to how data are modeled, processed, and interpreted. These include the development of hybrid and physics-informed models capable of handling uncertainty \cite{guo_combined_2022,coraddu_vessels_2017, ma2023interpretable}, the integration of multi-source data fusion frameworks \cite{du_data_2022, wang2023data}, and the creation of online or incremental learning approaches suitable for streaming conditions \cite{kaklis_online_2022, yuan_prediction_2021}. Most existing models are trained offline and struggle to adapt to changing operational or environmental contexts \cite{chi2018framework}.
    
Furthermore, most of the data-driven models for FOC are not explainable, which is essentially a challenge. Recently, various frameworks have been employed alongside these methods to incorporate explainability. For instance, in \cite{ma2023interpretable}, the SHAP framework \cite{lundberg2017unified} is exploited to interpret and analyze the relative importance of various influencing factors on FOC. \cite{wang2023innovative} proposed two ways to address the trade-off between interpretability and accuracy for FOC. The first approach uses a physics-informed neural network that combines domain knowledge with Bayesian belief propagation to enhance interpretability without compromising accuracy. The second approach utilizes the mixed-integer quadratic optimization model, which is an explainable Bayesian belief model obtained by solving an MIO model. Comparing the performance of the PI-NN and MIO models reveals that the PI-NN is better suited to high-accuracy cases, whereas the MIO is better suited to high-interpretability cases. Additionally, ensuring interpretability and transparency through XAI remains a key issue for stakeholder trust and regulatory compliance \cite{wang2023innovative, ma2023interpretable, wang2023data, ma2023multi}.

\subsubsection{Operational: Online Optimization, Deployment, and Real-time Monitoring}

Most available optimization methods for fuel consumption optimization are batch. In reality, meteorological conditions and other factors change in real time. Consequently, it is imperative to develop real-time optimization methods that incorporate the latest information, process it effectively, and produce outputs that facilitate immediate decision-making. In addition, for such complex online scenarios, factors such as the selection of the optimization variable, the optimality of the solution, and the computational complexity of these methods should also be investigated.

Beyond algorithmic performance, several operational constraints affect real-world deployment. These include the lack of onboard computational resources for real-time inference \cite{chi2018framework,chi2015ais}, variability across vessel types \cite{merien2018situ}, and integration of AI-driven recommendations into existing decision-support systems \cite{gkerekos_novel_2020,vettor2015multi}. Developing scalable, explainable, and human-in-the-loop frameworks for real-time monitoring and optimization is still an open challenge \cite{ma2023interpretable, wang_joint_2023, chi2018framework}. Moreover, aligning technological advances with regulatory and economic realities, such as carbon-intensity metrics, fuel pricing, and emission-control policies, requires interdisciplinary collaboration \cite{luo2023after,ma_method_2020,zhao_bi-objective_2019}.

\subsection{Limitations}
The existing literature contains certain limitations that hinder its practical applications and usefulness. 
\begin{itemize}
    \item Data Fusion. There are various sources of data related to FOC. Although works are available in the literature for data fusion, the available data from multiple sources are generally not used to estimate FOC by fusion. 
    \item No Loading Information. Some works do not include ship loading information, such as draft, loading status (ballast or laden), cargo weight (in tons), and other relevant details. As loading conditions affect fuel consumption, excluding them from the analysis as a feature limits the application of the FOC estimation model. 
    \item Relative Quantities. Some meteorological factors, such as wind and sea currents, should not be considered directly in the analysis. Rather, a relative quantity that considers the vessel's course can be computed. 
    \item Generic Models. Generic models for FOC estimation and optimization applicable to all kinds of vessels are very rare in the maritime literature. This is why these works rarely compare the proposed algorithms with the state-of-the-art ones. Most works are specific to certain types of ships, data types, sailing conditions, fuel types, and other relevant factors.    
\end{itemize}

\section{Conclusion and Future Direction}\label{sec:conclusion}
We have reviewed the FOC estimation and optimization models available in the literature. We have also discussed the challenges of FOC estimation and optimization. Moreover, the common limitations of the available works on FOC estimation and optimization are also discussed in detail. Additionally, future directions for research on fuel oil consumption estimation and optimization are outlined. There has also been a shift from conventional shipping to alternative technologies, such as green shipping. The goal is to develop green technologies for shipping that reduce emissions and lower costs. For example, multi-energy hybrid propulsion systems are emerging as a vital innovation for the future of maritime transport as discussed in \cite{guo2024energy}. In an all-electric ship (AES), considering uncertainties in the navigation environment and load demand, a joint optimization for power generation and voyage scheduling is formulated and solved, to minimize operating and battery loss costs, using reinforcement learning in \cite{shang2024dynamic}.

Furthermore, there are economic aspects related to fuel consumption, such as high bunker prices or crises in the maritime market, e.g., demand drop or fleet oversupply. In these cases, as fuel consumption rises with the cube of speed, to reduce costs, ship operators often choose slow steaming. i.e., voluntary speed reduction by ships as an effective measure to reduce fuel consumption and operational costs.

Finally, we highlight future research directions in FOC estimation and optimization as follows.\\

\textbf{Future Directions:}
\begin{itemize}
    \item An important research direction in the FOC estimation is the fusion of different data sources. Although some approaches to data fusion are available, there is still a need to expand the list of state-of-the-art techniques for data fusion.
    
    \item There should be a publicly available standard dataset for FOC, which contains data from various data sources for all major types of vessels. It will help researchers in the field to create benchmark algorithms and compare different data-driven algorithms.
    
    \item Different machine learning algorithms are mentioned in this review paper; however, the same techniques with different hyperparameters perform differently. For example, the same technique works perfectly in one work, but not in another. Therefore, all details related to the data used, preprocessing methods, model hyperparameters, and other important information should be mentioned to reproduce the results and clarify them for the reader. Thus, in future work, there is significant potential to propose FOC estimation and optimization methods that consider these issues.
    
    \item Hybrid models for FOC estimation have not been well explored in the literature. For example, modifying the loss function to incorporate information about the ship's underlying features could be investigated.

    \item Online algorithms for estimating the FOC of ships may also be explored further. For instance, the FOC estimation algorithm can detect the mode of the ship and apply the corresponding model in real time according to the ship's activity.

    \item Future work should focus on developing real-time optimization frameworks that integrate various data streams (e.g., AIS, weather forecasts, onboard sensors) and process them dynamically. Furthermore, to improve model generalization, dataset standardization across different vessel types should be prioritized. Collaborations with industry stakeholders could provide the data necessary for such efforts, as well as for validating the effectiveness of hybrid models.
    
    \item Future research should explore hybrid XAI approaches on how PI-NN, SHAP, and BNN can be combined to improve both interpretability and accuracy in FOC estimation.
    
    \item Developing XAI-driven real-time FOC monitoring can improve fuel efficiency and environmental compliance.
    
    \item A standard dataset for benchmarking explainable AI methods should be created to evaluate the trade-offs between black-box deep learning models and interpretable XAI models.
\end{itemize}

% To print the credit authorship contribution details
%\printcredits
\section*{Acknowledgment}
This work is supported by the GASS project funded by the Research Council of Norway under grant agreement No. 346603.
\section*{Data Statement}
No data was used in this article.

\bibliographystyle{elsarticle-num-names} 
\bibliography{cas-refs}

@article{bayraktar2024marine,
  title={Marine vessel energy efficiency performance prediction based on daily reported noon reports},
  author={Bayraktar, Murat and Sokukcu, Mustafa},
  journal={Ships and Offshore Structures},
  volume={19},
  number={6},
  pages={831--840},
  year={2024},
  publisher={Taylor \& Francis}
}

@article{wang2023innovative,
  title={Innovative approaches to addressing the tradeoff between interpretability and accuracy in ship fuel consumption prediction},
  author={Wang, Haoqing and Yan, Ran and Wang, Shuaian and Zhen, Lu},
  journal={Transportation Research Part C: Emerging Technologies},
  volume={157},
  pages={104361},
  year={2023},
  publisher={Elsevier}
}

@article{farag2020development,
  title={The development of a ship performance model in varying operating conditions based on ANN and regression techniques},
  author={Farag, Yasser BA and {\"O}l{\c{c}}er, Aykut I},
  journal={Ocean Engineering},
  volume={198},
  pages={106972},
  year={2020},
  publisher={Elsevier}
}

@article{shu2024investigation,
  title={Investigation of ship energy consumption based on neural network},
  author={Shu, Yaqing and Yu, Benshuang and Liu, Wei and Yan, Tao and Liu, Zhiyao and Gan, Langxiong and Yin, Jianchuan and Song, Lan},
  journal={Ocean \& Coastal Management},
  volume={254},
  pages={107167},
  year={2024},
  publisher={Elsevier}
}

@article{veneti2017minimizing,
  title={Minimizing the fuel consumption and the risk in maritime transportation: A bi-objective weather routing approach},
  author={Veneti, Aphrodite and Makrygiorgos, Angelos and Konstantopoulos, Charalampos and Pantziou, Grammati and Vetsikas, Ioannis A},
  journal={Computers \& Operations Research},
  volume={88},
  pages={220--236},
  year={2017},
  publisher={Elsevier}
}

@article{shang2024dynamic,
  title={Dynamic joint optimization of power generation and voyage scheduling in ship power system based on deep reinforcement learning},
  author={Shang, Chengya and Fu, Lijun and Bao, Xianqiang and Xiao, Haipeng and Xu, Xinghua and Hu, Qi},
  journal={Electric Power Systems Research},
  volume={229},
  pages={110165},
  year={2024},
  publisher={Elsevier}
}

@article{guo2024energy,
  title={Energy management system for hybrid ship: Status and perspectives},
  author={Guo, Xiaodong and Lang, Xiao and Yuan, Yupeng and Tong, Liang and Shen, Boyang and Long, Teng and Mao, Wengang},
  journal={Ocean Engineering},
  volume={310},
  pages={118638},
  year={2024},
  publisher={Elsevier}
}

@article{mylonopoulos2023comprehensive,
  title={A comprehensive review of modeling and optimization methods for ship energy systems},
  author={Mylonopoulos, Foivos and Polinder, Henk and Coraddu, Andrea},
  journal={IEEE Access},
  volume={11},
  pages={32697--32707},
  year={2023},
  publisher={IEEE}
}

@article{rudzki2022optimization,
  title={Optimization model to manage ship fuel consumption and navigation time},
  author={Rudzki, Krzysztof and Gomulka, Piotr and Hoang, Anh Tuan},
  journal={Polish Maritime Research},
  volume={29},
  number={3},
  pages={141--153},
  year={2022}
}

@article{yu2024trim,
  title={Trim and Engine Power Joint Optimization of a Ship Based on Minimum Energy Consumption over a Whole Voyage},
  author={Yu, Yanyun and Zhang, Hongshuo and Mu, Zongbao and Li, Yating and Sun, Yutong and Liu, Jia},
  journal={Journal of Marine Science and Engineering},
  volume={12},
  number={3},
  pages={475},
  year={2024},
  publisher={MDPI}
}

@article{rai2020driven,
  title={Driven by data or derived through physics? a review of hybrid physics guided machine learning techniques with cyber-physical system (cps) focus},
  author={Rai, Rahul and Sahu, Chandan K},
  journal={IEEE Access},
  volume={8},
  pages={71050--71073},
  year={2020},
  publisher={IEEE}
}

@inproceedings{liu2020voyage,
  title={Voyage performance evaluation based on a digital twin model},
  author={Liu, M and Zhou, Q and Wang, X and Yu, C and Kang, M},
  booktitle={IOP Conference Series: Materials Science and Engineering},
  volume={929},
  number={1},
  pages={012027},
  year={2020},
  organization={IOP Publishing}
}

@article{ruan2024novel,
  title={A novel prediction method of fuel consumption for wing-diesel hybrid vessels based on feature construction},
  author={Ruan, Zhang and Huang, Lianzhong and Wang, Kai and Ma, Ranqi and Wang, Zhongyi and Zhang, Rui and Zhao, Haoyang and Wang, Cong},
  journal={Energy},
  volume={286},
  pages={129516},
  year={2024},
  publisher={Elsevier}
}

@article{lundberg2017unified,
  title={A unified approach to interpreting model predictions},
  author={Lundberg, Scott M and Lee, Su-In},
  journal={Advances in neural information processing systems},
  volume={30},
  year={2017}
}

@article{ma2023interpretable,
  title={An interpretable gray box model for ship fuel consumption prediction based on the SHAP framework},
  author={Ma, Yiji and Zhao, Yuzhe and Yu, Jiahao and Zhou, Jingmiao and Kuang, Haibo},
  journal={Journal of Marine Science and Engineering},
  volume={11},
  number={5},
  pages={1059},
  year={2023},
  publisher={MDPI}
}

@article{su2023fuel,
  title={Fuel Consumption Prediction and Optimization Model for Pure Car/Truck Transport Ships},
  author={Su, Miao and Su, Zhenqing and Cao, Shengli and Park, Keun-Sik and Bae, Sung-Hoon},
  journal={Journal of Marine Science and Engineering},
  volume={11},
  number={6},
  pages={1231},
  year={2023},
  publisher={MDPI}
}

@article{zeng2022data,
  title={A data-driven intelligent energy efficiency management system for ships},
  author={Zeng, Xiangming and Chen, Mingzhi and Li, Hongfei and Wu, Xianhua},
  journal={IEEE Intelligent Transportation Systems Magazine},
  volume={15},
  number={1},
  pages={270--284},
  year={2022},
  publisher={IEEE}
}

@article{yan2018energy,
  title={Energy-efficient shipping: An application of big data analysis for optimizing engine speed of inland ships considering multiple environmental factors},
  author={Yan, Xinping and Wang, Kai and Yuan, Yupeng and Jiang, Xiaoli and Negenborn, Rudy R},
  journal={Ocean Engineering},
  volume={169},
  pages={457--468},
  year={2018},
  publisher={Elsevier}
}

@article{wang2016real,
  title={Real-time optimization of ship energy efficiency based on the prediction technology of working condition},
  author={Wang, Kai and Yan, Xinping and Yuan, Yupeng and Li, Feng},
  journal={Transportation Research Part D: Transport and Environment},
  volume={46},
  pages={81--93},
  year={2016},
  publisher={Elsevier}
}

@article{wang2018predicting,
  title={Predicting ship fuel consumption based on LASSO regression},
  author={Wang, Shengzheng and Ji, Baoxian and Zhao, Jiansen and Liu, Wei and Xu, Tie},
  journal={Transportation Research Part D: Transport and Environment},
  volume={65},
  pages={817--824},
  year={2018},
  publisher={Elsevier}
}

@article{gkerekos2019machine,
  title={Machine learning models for predicting ship main engine Fuel Oil Consumption: A comparative study},
  author={Gkerekos, Christos and Lazakis, Iraklis and Theotokatos, Gerasimos},
  journal={Ocean Engineering},
  volume={188},
  pages={106282},
  year={2019},
  publisher={Elsevier}
}

@article{zakerdoost2019multi,
  title={A multi-level optimization technique based on fuel consumption and energy index in early-stage ship design},
  author={Zakerdoost, Hassan and Ghassemi, Hassan},
  journal={Structural and Multidisciplinary Optimization},
  volume={59},
  pages={1417--1438},
  year={2019},
  publisher={Springer}
}

@article{wang2023novel,
  title={Novel ship fuel consumption modelling approaches for speed and trim optimisation: Using engine data as auxiliary},
  author={Wang, Kangli and Zhang, Defu and Shen, Zhenyu and Zhu, Wei and Ye, Hongcai and Li, Dong},
  journal={Ocean Engineering},
  volume={286},
  pages={115520},
  year={2023},
  publisher={Elsevier}
}

@article{xie2023joint,
  title={Joint optimization of ship speed and trim based on machine learning method under consideration of load},
  author={Xie, Xianwei and Sun, Baozhi and Li, Xiaohe and Zhao, Yuhao and Chen, Yumei},
  journal={Ocean Engineering},
  volume={287},
  pages={115917},
  year={2023},
  publisher={Elsevier}
}

@article{fan2022joint,
  title={Joint optimisation for improving ship energy efficiency considering speed and trim control},
  author={Fan, Ailong and Yang, Jian and Yang, Liu and Liu, Weiqin and Vladimir, Nikola},
  journal={Transportation Research Part D: Transport and Environment},
  volume={113},
  pages={103527},
  year={2022},
  publisher={Elsevier}
}

@article{islam2019effect,
  title={Effect of trim on container ship resistance at different ship speeds and drafts},
  author={Islam, Hafizul and Soares, Carlos Guedes},
  journal={Ocean Engineering},
  volume={183},
  pages={106--115},
  year={2019},
  publisher={Elsevier}
}

@article{reichel2014trim,
  title={Trim optimisation-theory and practice},
  author={Reichel, Maciej and Minchev, Anton and Larsen, Nikolaj Lemb},
  journal={TransNav: International Journal on Marine Navigation and Safety of Sea Transportation},
  volume={8},
  number={3},
  year={2014}
}

@article{kaklis2023enabling,
  title={Enabling digital twins in the maritime sector through the lens of AI and industry 4.0},
  author={Kaklis, Dimitrios and Varlamis, Iraklis and Giannakopoulos, George and Varelas, Takis J and Spyropoulos, Constantine D},
  journal={International Journal of Information Management Data Insights},
  volume={3},
  number={2},
  pages={100178},
  year={2023},
  publisher={Elsevier}
}

@article{yan2020development,
  title={Development of a two-stage ship fuel consumption prediction and reduction model for a dry bulk ship},
  author={Yan, Ran and Wang, Shuaian and Du, Yuquan},
  journal={Transportation Research Part E: Logistics and Transportation Review},
  volume={138},
  pages={101930},
  year={2020},
  publisher={Elsevier}
}

@article{kee2018prediction,
  title={Prediction of ship fuel consumption and speed curve by using statistical method},
  author={Kee, Keh-Kim and Simon, B-Y Lau and Renco, K-H Yong},
  journal={J. Comput. Sci. Comput. Math},
  volume={8},
  number={2},
  pages={19--24},
  year={2018}
}

@article{gao2021speed,
  title={Speed optimization for container ship fleet deployment considering fuel consumption},
  author={Gao, Chao-Feng and Hu, Zhi-Hua},
  journal={Sustainability},
  volume={13},
  number={9},
  pages={5242},
  year={2021},
  publisher={MDPI}
}

@article{taskar2023case,
  title={A case study for the assessment of fuel savings using speed optimization},
  author={Taskar, Bhushan and Sasmal, Kaushik and Yiew, Lucas J},
  journal={Ocean Engineering},
  volume={274},
  pages={113990},
  year={2023},
  publisher={Elsevier}
}

@article{medina2020bunker,
  title={Bunker consumption of containerships considering sailing speed and wind conditions},
  author={Medina, Josep R and Molines, Jorge and Gonz{\'a}lez-Escriv{\'a}, Jos{\'e} A and Aguilar, Jos{\'e}},
  journal={Transportation Research Part D: Transport and Environment},
  volume={87},
  pages={102494},
  year={2020},
  publisher={Elsevier}
}

@article{yang2020ship,
  title={Ship speed optimization considering ocean currents to enhance environmental sustainability in maritime shipping},
  author={Yang, Liqian and Chen, Gang and Zhao, Jinlou and Rytter, Niels Gorm Mal{\`y}},
  journal={Sustainability},
  volume={12},
  number={9},
  pages={3649},
  year={2020},
  publisher={MDPI}
}

@article{sasa2021speed,
  title={Speed loss analysis and rough wave avoidance algorithms for optimal ship routing simulation of 28,000-DWT bulk carrier},
  author={Sasa, Kenji and Chen, Chen and Fujimatsu, Takuya and Shoji, Ruri and Maki, Atsuo},
  journal={Ocean Engineering},
  volume={228},
  pages={108800},
  year={2021},
  publisher={Elsevier}
}

@article{ormevik2023high,
  title={A high-fidelity approach to modeling weather-dependent fuel consumption on ship routes with speed optimization},
  author={Ormevik, Andreas Breivik and Fagerholt, Kjetil and Meisel, Frank and Sandvik, Endre},
  journal={Maritime Transport Research},
  volume={5},
  pages={100096},
  year={2023},
  publisher={Elsevier}
}

@article{du2021estimation,
  title={Estimation of ship routes considering weather and constraints},
  author={Du, Wei and Li, Yanjun and Zhang, Guolei and Wang, Chunhui and Chen, Pan and Qiao, Jipan},
  journal={Ocean Engineering},
  volume={228},
  pages={108695},
  year={2021},
  publisher={Elsevier}
}

@article{godet2023prediction,
  title={Prediction of container ships’ speed-power relationship for various operational conditions},
  author={Godet, Amandine},
  journal={Transportation Research Procedia},
  volume={72},
  pages={1861--1868},
  year={2023},
  publisher={Elsevier}
}

@article{tillig2018analysis,
  title={Analysis of uncertainties in the prediction of ships’ fuel consumption--from early design to operation conditions},
  author={Tillig, Fabian and Ringsberg, Jonas W and Mao, Wengang and Ramne, Bengt},
  journal={Ships and Offshore Structures},
  volume={13},
  number={sup1},
  pages={13--24},
  year={2018},
  publisher={Taylor \& Francis}
}

@article{berthelsen2021prediction,
  title={Prediction of ships’ speed-power relationship at speed intervals below the design speed},
  author={Berthelsen, Frederik H and Nielsen, Ulrik D},
  journal={Transportation Research Part D: Transport and Environment},
  volume={99},
  pages={102996},
  year={2021},
  publisher={Elsevier}
}

@inproceedings{tvete2020modelling,
  title={A modelling system for power consumption of marine traffic},
  author={Tvete, Hans Anton and Guo, Bingjie and Liang, Qin and Brinks, Hendrik},
  booktitle={International Conference on Offshore Mechanics and Arctic Engineering},
  volume={84379},
  pages={V06AT06A029},
  year={2020},
  organization={American Society of Mechanical Engineers}
}

@article{elkafas2022advanced,
  title={Advanced operational measure for reducing fuel consumption onboard ships},
  author={Elkafas, Ahmed G},
  journal={Environmental Science and Pollution Research},
  volume={29},
  number={60},
  pages={90509--90519},
  year={2022},
  publisher={Springer}
}

@incollection{parkes2019efficient,
  title={Efficient vessel power prediction in operational conditions using machine learning},
  author={Parkes, AI and Savasta, TD and Sobey, AJ and Hudson, DA},
  booktitle={Practical Design of Ships and Other Floating Structures},
  pages={350--367},
  year={2019},
  publisher={Springer}
}

@article{gunecs2023predicting,
  title={Predicting tanker main engine power using regression analysis and artificial neural networks},
  author={G{\"u}ne{\c{s}}, {\"U}mit and Ba{\c{s}}han, Veysi and Karakurt, As{\i}m Sinan},
  journal={Sigma Journal of Engineering and Natural Sciences},
  volume={41},
  number={2},
  pages={216--225},
  year={2023},
  publisher={Yildiz Technical University}
}

@article{zhang2023development,
  title={Development of a ship performance model for power estimation of inland waterway vessels},
  author={Zhang, Chengqian and Ringsberg, Jonas W and Thies, Fabian},
  journal={Ocean Engineering},
  volume={287},
  pages={115731},
  year={2023},
  publisher={Elsevier}
}

@inproceedings{liang2019prediction,
  title={Prediction of vessel propulsion power using machine learning on AIS data, ship performance measurements and weather data},
  author={Liang, Qin and Tvete, Hans Anton and Brinks, Hendrik W},
  booktitle={Journal of physics: Conference series},
  volume={1357},
  number={1},
  pages={012038},
  year={2019},
  organization={IOP Publishing}
}

@article{laurie2021machine,
  title={Machine learning for shaft power prediction and analysis of fouling related performance deterioration},
  author={Laurie, Anastasia and Anderlini, Enrico and Dietz, Jesper and Thomas, Giles},
  journal={Ocean Engineering},
  volume={234},
  pages={108886},
  year={2021},
  publisher={Elsevier}
}

@article{lang2022comparison,
  title={Comparison of supervised machine learning methods to predict ship propulsion power at sea},
  author={Lang, Xiao and Wu, Da and Mao, Wengang},
  journal={Ocean Engineering},
  volume={245},
  pages={110387},
  year={2022},
  publisher={Elsevier}
}

@article{parkes2018physics,
  title={Physics-based shaft power prediction for large merchant ships using neural networks},
  author={Parkes, AI and Sobey, AJ and Hudson, DA},
  journal={Ocean Engineering},
  volume={166},
  pages={92--104},
  year={2018},
  publisher={Elsevier}
}

@article{lang2020semi,
  title={A semi-empirical model for ship speed loss prediction at head sea and its validation by full-scale measurements},
  author={Lang, Xiao and Mao, Wengang},
  journal={Ocean Engineering},
  volume={209},
  pages={107494},
  year={2020},
  publisher={Elsevier}
}

@article{lang2021practical,
  title={A practical speed loss prediction model at arbitrary wave heading for ship voyage optimization},
  author={Lang, Xiao and Mao, Wengang},
  journal={Journal of Marine Science and Application},
  volume={20},
  number={3},
  pages={410--425},
  year={2021},
  publisher={Springer}
}

@article{dekeyser2022towards,
  title={Towards improved prediction of ship performance: a comparative analysis on in-service ship monitoring data for modeling the speed-power relation},
  author={DeKeyser, Simon and Morob{\'e}, Casimir and Mittendorf, Malte},
  journal={arXiv preprint arXiv:2212.13061},
  year={2022}
}

@article{bialystocki2016estimation,
  title={On the estimation of ship's fuel consumption and speed curve: A statistical approach},
  author={Bialystocki, Nicolas and Konovessis, Dimitris},
  journal={Journal of Ocean Engineering and Science},
  volume={1},
  number={2},
  pages={157--166},
  year={2016},
  publisher={Elsevier}
}

@article{adland2020optimal,
  title={Optimal ship speed and the cubic law revisited: Empirical evidence from an oil tanker fleet},
  author={Adland, Roar and Cariou, Pierre and Wolff, Francois-Charles},
  journal={Transportation Research Part E: Logistics and Transportation Review},
  volume={140},
  pages={101972},
  year={2020},
  publisher={Elsevier}
}

@article{odendaal2023enhancing,
  title={Enhancing early-stage energy consumption predictions using dynamic operational voyage data: A grey-box modelling investigation},
  author={Odendaal, Kirsten and Alkemade, Aaron and Kana, Austin A},
  journal={International Journal of Naval Architecture and Ocean Engineering},
  volume={15},
  pages={100484},
  year={2023},
  publisher={Elsevier}
}

@article{yang2019genetic,
  title={A genetic algorithm-based grey-box model for ship fuel consumption prediction towards sustainable shipping},
  author={Yang, Liqian and Chen, Gang and Rytter, Niels Gorm Mal{\`y} and Zhao, Jinlou and Yang, Dong},
  journal={Annals of Operations Research},
  pages={1--27},
  year={2019},
  publisher={Springer}
}

@article{yan2024improving,
  title={Improving ship energy efficiency: Models, methods, and applications},
  author={Yan, Ran and Yang, Dong and Wang, Tianyu and Mo, Haoyu and Wang, Shuaian},
  journal={Applied Energy},
  volume={368},
  pages={123132},
  year={2024},
  publisher={Elsevier}
}

@article{papandreou2022predicting,
  title={Predicting VLCC fuel consumption with machine learning using operationally available sensor data},
  author={Papandreou, Christos and Ziakopoulos, Apostolos},
  journal={Ocean Engineering},
  volume={243},
  pages={110321},
  year={2022},
  publisher={Elsevier}
}

@article{zhou2023predicting,
  title={Predicting ship fuel consumption using a combination of metocean and on-board data},
  author={Zhou, Yi and Pazouki, Kayvan and Murphy, Alan J and Uriondo, Zigor and Granado, Igor and Quincoces, I{\~n}aki and Fernandes-Salvador, Jose A},
  journal={Ocean Engineering},
  volume={285},
  pages={115509},
  year={2023},
  publisher={Elsevier}
}

@article{merien2018situ,
  title={In-situ data vs. bottom-up approaches in estimations of marine fuel consumptions and emissions},
  author={Merien-Paul, Rumesh H and Enshaei, Hossein and Jayasinghe, Shantha Gamini},
  journal={Transportation Research Part D: Transport and Environment},
  volume={62},
  pages={619--632},
  year={2018},
  publisher={Elsevier}
}

@article{luo2023comparison,
  title={Comparison of deterministic and ensemble weather forecasts on ship sailing speed optimization},
  author={Luo, Xi and Yan, Ran and Wang, Shuaian},
  journal={Transportation Research Part D: Transport and Environment},
  volume={121},
  pages={103801},
  year={2023},
  publisher={Elsevier}
}

@article{safaei2019vlcc,
  title={VLCC’s fuel consumption prediction modeling based on noon report and automatic identification system},
  author={Safaei, Ali Akbar and Ghassemi, Hassan and Ghiasi, Mahmoud},
  journal={Cogent Engineering},
  volume={6},
  number={1},
  pages={1595292},
  year={2019},
  publisher={Taylor \& Francis}
}

@inproceedings{lu2013voyage,
  title={Voyage optimisation: prediction of ship specific fuel consumption for energy efficient shipping},
  author={Lu, Ruihua and Turan, Osman and Boulougouris, Evangelos},
  booktitle={3rd International Conference onTechnologies, Operations, Logistics and Modelling for Low Carbon Shipping},
  pages={1--11},
  year={2013}
}

@article{yan2021emerging,
  title={Emerging approaches applied to maritime transport research: Past and future},
  author={Yan, Ran and Wang, Shuaian and Zhen, Lu and Laporte, Gilbert},
  journal={Communications in Transportation Research},
  volume={1},
  pages={100011},
  year={2021},
  publisher={Elsevier}
}

@article{icsikli2020estimating,
  title={Estimating fuel consumption in maritime transport},
  author={I{\c{s}}{\i}kl{\i}, Erkan and Ayd{\i}n, Nezir and Bilgili, Levent and Toprak, Ali},
  journal={Journal of Cleaner Production},
  volume={275},
  pages={124142},
  year={2020},
  publisher={Elsevier}
}

@article{zwart2023grey,
  title={A Grey-box model approach using noon report data for trim optimization},
  author={Zwart, Robert H and Bogaard, Jordi and Kana, Austin A},
  journal={International Shipbuilding Progress},
  volume={70},
  number={1},
  pages={41--63},
  year={2023},
  publisher={IOS Press}
}

@article{mepc20232023,
  title={2023 IMO strategy on reduction of GHG emissions from ships},
  author={MEPC, RESOLUTION},
  year={2023}
}

@article{heikkila2024effect,
  title={Effect of ice class to vessel fuel consumption based on real-life MRV data},
  author={Heikkil{\"a}, Mikko and Gr{\"o}nholm, Tiia and Majam{\"a}ki, Elisa and Jalkanen, Jukka-Pekka},
  journal={Transport Policy},
  volume={148},
  pages={168--180},
  year={2024},
  publisher={Elsevier}
}

@article{zis2020ship,
  title={Ship weather routing: A taxonomy and survey},
  author={Zis, Thalis PV and Psaraftis, Harilaos N and Ding, Li},
  journal={Ocean Engineering},
  volume={213},
  pages={107697},
  year={2020},
  publisher={Elsevier}
}

@article{vettor2015multi,
  title={Multi-objective route optimization for onboard decision support system},
  author={Vettor, R and Guedes Soares, C},
  journal={Information, Communication and Environment: Marine Navigation and Safety of Sea Transportation; Weintrit, A., Neumann, T., Eds},
  pages={99--106},
  year={2015}
}

@article{doundoulakis2022comparative,
  title={Comparative analysis of fuel consumption and CO2 emission estimation based on ships activity and reported fuel consumption: the case of short sea shipping in Crete},
  author={Doundoulakis, Emmanouil and Papaefthimiou, Spiros},
  journal={Greenhouse Gases: Science and Technology},
  volume={12},
  number={5},
  pages={629--641},
  year={2022},
  publisher={Wiley Online Library}
}

@article{hu2022two,
  title={A two-step strategy for fuel consumption prediction and optimization of ocean-going ships},
  author={Hu, Zhihui and Zhou, Tianrui and Zhen, Rong and Jin, Yongxing and Li, Xiaohe and Osman, Mohd Tarmizi},
  journal={Ocean Engineering},
  volume={249},
  pages={110904},
  year={2022},
  publisher={Elsevier}
}

@article{ma2023multi,
  title={A multi-objective energy efficiency optimization method of ship under different sea conditions},
  author={Ma, Lin and Yang, Ping and Gao, Diju and Bao, Chunteng},
  journal={Ocean Engineering},
  volume={290},
  pages={116337},
  year={2023},
  publisher={Elsevier}
}

@article{luo2023after,
  title={After five years’ application of the European Union monitoring, reporting, and verification (MRV) mechanism: Review and prospectives},
  author={Luo, Xi and Yan, Ran and Wang, Shuaian},
  journal={Journal of Cleaner Production},
  pages={140006},
  year={2023},
  publisher={Elsevier}
}

@article{yan2023analysis,
  title={Analysis and prediction of ship energy efficiency based on the MRV system},
  author={Yan, Ran and Mo, Haoyu and Wang, Shuaian and Yang, Dong},
  journal={Maritime Policy \& Management},
  volume={50},
  number={1},
  pages={117--139},
  year={2023},
  publisher={Taylor \& Francis}
}

@article{moreira2021neural,
  title={Neural network approach for predicting ship speed and fuel consumption},
  author={Moreira, L{\'u}cia and Vettor, Roberto and Guedes Soares, Carlos},
  journal={Journal of Marine Science and Engineering},
  volume={9},
  number={2},
  pages={119},
  year={2021},
  publisher={Multidisciplinary Digital Publishing Institute}
}

@inproceedings{chi2015ais,
  title={An AIS-based framework for real time monitoring of vessels efficiency},
  author={Chi, Hongtao and Pedrielli, Giulia and Kister, Thomas and Ng, Szu Hui and Bressan, St{\'e}phane},
  booktitle={2015 IEEE International Conference on Industrial Engineering and Engineering Management (IEEM)},
  pages={1218--1222},
  year={2015},
  organization={IEEE}
}

@article{yan2021data,
  title={Data analytics for fuel consumption management in maritime transportation: Status and perspectives},
  author={Yan, Ran and Wang, Shuaian and Psaraftis, Harilaos N},
  journal={Transportation Research Part E: Logistics and Transportation Review},
  volume={155},
  pages={102489},
  year={2021},
  publisher={Elsevier}
}

@article{chi2018framework,
  title={A framework for real-time monitoring of energy efficiency of marine vessels},
  author={Chi, Hongtao and Pedrielli, Giulia and Ng, Szu Hui and Kister, Thomas and Bressan, St{\'e}phane},
  journal={Energy},
  volume={145},
  pages={246--260},
  year={2018},
  publisher={Elsevier}
}

@article{kim2021development,
  title={Development of a fuel consumption prediction model based on machine learning using ship in-service data},
  author={Kim, Young-Rong and Jung, Min and Park, Jun-Bum},
  journal={Journal of Marine Science and Engineering},
  volume={9},
  number={2},
  pages={137},
  year={2021},
  publisher={Multidisciplinary Digital Publishing Institute}
}

@article{zhu2021modeling,
  title={Modeling of ship fuel consumption based on multisource and heterogeneous data: Case study of passenger ship},
  author={Zhu, Yongjie and Zuo, Yi and Li, Tieshan},
  journal={Journal of Marine Science and Engineering},
  volume={9},
  number={3},
  pages={273},
  year={2021},
  publisher={MDPI}
}

@article{zhang2023research,
  title={Research on Carbon Intensity Prediction Method for Ships Based on Sensors and Meteorological Data},
  author={Zhang, Chunchang and Lu, Tianye and Wang, Zhihuan and Zeng, Xiangming},
  journal={Journal of Marine Science and Engineering},
  volume={11},
  number={12},
  pages={2249},
  year={2023},
  publisher={MDPI}
}

@article{wang2023ship,
  title={Ship fuel and carbon emission estimation utilizing artificial neural network and data fusion techniques},
  author={Wang, Shaohan and Wang, Xinbo and Han, Yi and Wang, Xiangyu and Jiang, He and Zhang, Zhexi},
  journal={Journal of Software Engineering and Applications},
  volume={16},
  number={3},
  pages={51--72},
  year={2023},
  publisher={Scientific Research Publishing}
}

@article{ren2022container,
  title={Container ship carbon and fuel estimation in voyages utilizing meteorological data with data fusion and machine learning techniques},
  author={Ren, Feiyang and Wang, Shaohan and Liu, Yuanzhe and Han, Yi},
  journal={Mathematical Problems in Engineering},
  volume={2022},
  number={1},
  pages={4773395},
  year={2022},
  publisher={Wiley Online Library}
}

@article{karagiannidis2021data,
  title={Data-driven modelling of ship propulsion and the effect of data pre-processing on the prediction of ship fuel consumption and speed loss},
  author={Karagiannidis, Pavlos and Themelis, Nikos},
  journal={Ocean Engineering},
  volume={222},
  pages={108616},
  year={2021},
  publisher={Elsevier}
}

@article{coraddu_data-driven_2019,
	title = {Data-driven ship digital twin for estimating the speed loss caused by the marine fouling},
	volume = {186},
	issn = {00298018},
	doi = {10.1016/j.oceaneng.2019.05.045},
	language = {en},
	urldate = {2024-01-18},
	journal = {Ocean Engineering},
	author = {Coraddu, Andrea and Oneto, Luca and Baldi, Francesco and Cipollini, Francesca and Atlar, Mehmet and Savio, Stefano},
	month = aug,
	year = {2019},
	pages = {106063},
}

@article{yan_data_2021,
	title = {Data analytics for fuel consumption management in maritime transportation: {Status} and perspectives},
	volume = {155},
	issn = {13665545},
	shorttitle = {Data analytics for fuel consumption management in maritime transportation},
	doi = {10.1016/j.tre.2021.102489},
	language = {en},
	urldate = {2024-03-05},
	journal = {Transportation Research Part E: Logistics and Transportation Review},
	author = {Yan, Ran and Wang, Shuaian and Psaraftis, Harilaos N.},
	month = nov,
	year = {2021},
	pages = {102489},
}

@article{wang_joint_2023,
	title = {Joint {Voyage} {Planning} and {Onboard} {Energy} {Management} of {Hybrid} {Propulsion} {Ships}},
	volume = {11},
	issn = {2077-1312},
	doi = {10.3390/jmse11030585},
	number = {3},
	urldate = {2024-03-05},
	journal = {Journal of Marine Science and Engineering},
	author = {Wang, Yu and Liang, Chengji and Aktas, Tugce Uslu and Shi, Jian and Pan, Yang and Fang, Sidun and Lim, Gino},
	month = mar,
	year = {2023},
	pages = {585},
}

@article{zhang_deep_2024,
	title = {A deep learning method for the prediction of ship fuel consumption in real operational conditions},
	volume = {130},
	issn = {09521976},
	doi = {10.1016/j.engappai.2023.107425},
	language = {en},
	journal = {Engineering Applications of Artificial Intelligence},
	author = {Zhang, Mingyang and Tsoulakos, Nikolaos and Kujala, Pentti and Hirdaris, Spyros},
	month = apr,
	year = {2024},
	pages = {107425},
}

@article{zheng_voyage_2019,
	title = {A voyage with minimal fuel consumption for cruise ships},
	volume = {215},
	issn = {09596526},
	doi = {10.1016/j.jclepro.2019.01.032},
	language = {en},
	urldate = {2024-03-05},
	journal = {Journal of Cleaner Production},
	author = {Zheng, Jianqin and Zhang, Haoran and Yin, Long and Liang, Yongtu and Wang, Bohong and Li, Zhengbing and Song, Xuan and Zhang, Yu},
	month = apr,
	year = {2019},
	pages = {144--153},
}

@article{zhao_bi-objective_2019,
	title = {Bi-{Objective} {Optimization} of {Vessel} {Speed} and {Route} for {Sustainable} {Coastal} {Shipping} under the {Regulations} of {Emission} {Control} {Areas}},
	volume = {11},
	issn = {2071-1050},
	doi = {10.3390/su11226281},
	language = {en},
	number = {22},
	urldate = {2024-03-05},
	journal = {Sustainability},
	author = {Zhao, Yuzhe and Fan, Yujun and Zhou, Jingmiao and Kuang, Haibo},
	month = nov,
	year = {2019},
	pages = {6281},
}

@article{fan_comprehensive_2024,
	title = {Comprehensive evaluation of machine learning models for predicting ship energy consumption based on onboard sensor data},
	volume = {248},
	issn = {09645691},
	doi = {10.1016/j.ocecoaman.2023.106946},
	language = {en},
	urldate = {2024-03-05},
	journal = {Ocean \& Coastal Management},
	author = {Fan, Ailong and Wang, Yingqi and Yang, Liu and Tu, Xiaolong and Yang, Jian and Shu, Yaqing},
	month = feb,
	year = {2024},
	pages = {106946},
}

@article{du_energy_2022,
	title = {Energy saving method for ship weather routing optimization},
	volume = {258},
	issn = {00298018},
	doi = {10.1016/j.oceaneng.2022.111771},
	language = {en},
	urldate = {2024-03-05},
	journal = {Ocean Engineering},
	author = {Du, Wei and Li, Yanjun and Zhang, Guolei and Wang, Chunhui and Zhu, Baitong and Qiao, Jipan},
	month = aug,
	year = {2022},
	pages = {111771},
}

@article{yu_literature_2021,
	title = {Literature review on emission control-based ship voyage optimization},
	volume = {93},
	issn = {13619209},
	doi = {10.1016/j.trd.2021.102768},
	language = {en},
	urldate = {2024-03-05},
	journal = {Transportation Research Part D: Transport and Environment},
	author = {Yu, Hongchu and Fang, Zhixiang and Fu, Xiuju and Liu, Jingxian and Chen, Jinhai},
	month = apr,
	year = {2021},
	pages = {102768},
}

@article{moradi_marine_2022,
	title = {Marine route optimization using reinforcement learning approach to reduce fuel consumption and consequently minimize {CO2} emissions},
	volume = {259},
	issn = {00298018},
	doi = {10.1016/j.oceaneng.2022.111882},
	language = {en},
	urldate = {2024-03-05},
	journal = {Ocean Engineering},
	author = {Moradi, Mohammad Hossein and Brutsche, Martin and Wenig, Markus and Wagner, Uwe and Koch, Thomas},
	month = sep,
	year = {2022},
	pages = {111882},
}

@article{ma_method_2020,
	title = {Method for simultaneously optimizing ship route and speed with emission control areas},
	volume = {202},
	issn = {00298018},
	doi = {10.1016/j.oceaneng.2020.107170},
	language = {en},
	urldate = {2024-03-05},
	journal = {Ocean Engineering},
	author = {Ma, Dongfang and Ma, Weihao and Jin, Sheng and Ma, Xiaolong},
	month = apr,
	year = {2020},
	pages = {107170},
}

@article{yuan_prediction_2021,
	title = {Prediction and optimisation of fuel consumption for inland ships considering real-time status and environmental factors},
	volume = {221},
	issn = {00298018},
	doi = {10.1016/j.oceaneng.2020.108530},
	language = {en},
	urldate = {2024-03-05},
	journal = {Ocean Engineering},
	author = {Yuan, Zhi and Liu, Jingxian and Zhang, Qian and Liu, Yi and Yuan, Yuan and Li, Zongzhi},
	month = feb,
	year = {2021},
	pages = {108530},
}

@article{hu_prediction_2019,
	title = {Prediction of {Fuel} {Consumption} for {Enroute} {Ship} {Based} on {Machine} {Learning}},
	volume = {7},
	issn = {2169-3536},
	doi = {10.1109/ACCESS.2019.2933630},
	language = {en},
	urldate = {2024-03-05},
	journal = {IEEE Access},
	author = {Hu, Zhihui and Jin, Yongxin and Hu, Qinyou and Sen, Sukanta and Zhou, Tianrui and Osman, Mohd Tarmizi},
	year = {2019},
	pages = {119497--119505},
}

@article{sang_ship_2023,
	title = {Ship voyage optimization based on fuel consumption under various operational conditions},
	volume = {352},
	issn = {00162361},
	doi = {10.1016/j.fuel.2023.129086},
	language = {en},
	urldate = {2024-03-05},
	journal = {Fuel},
	author = {Sang, Yijian and Ding, Yu and Xu, Jiarun and Sui, Congbiao},
	month = nov,
	year = {2023},
	pages = {129086},
}

@article{bui-duy_utilization_2021,
	title = {Utilization of a deep learning-based fuel consumption model in choosing a liner shipping route for container ships in {Asia}},
	volume = {37},
	issn = {20925212},
	doi = {10.1016/j.ajsl.2020.04.003},
	language = {en},
	number = {1},
	urldate = {2024-03-05},
	journal = {The Asian Journal of Shipping and Logistics},
	author = {Bui-Duy, Linh and Vu-Thi-Minh, Ngoc},
	month = mar,
	year = {2021},
	pages = {1--11},
}

@article{wang_voyage_2021,
	title = {Voyage optimization combining genetic algorithm and dynamic programming for fuel/emissions reduction},
	volume = {90},
	issn = {13619209},
	doi = {10.1016/j.trd.2020.102670},
	language = {en},
	urldate = {2024-03-05},
	journal = {Transportation Research Part D: Transport and Environment},
	author = {Wang, Helong and Lang, Xiao and Mao, Wengang},
	month = jan,
	year = {2021},
	pages = {102670},
}

@article{weng_ship_2020,
	title = {Ship emission estimation with high spatial-temporal resolution in the {Yangtze} {River} estuary using {AIS} data},
	volume = {248},
	issn = {09596526},
	doi = {10.1016/j.jclepro.2019.119297},
	language = {en},
	urldate = {2024-03-05},
	journal = {Journal of Cleaner Production},
	author = {Weng, Jinxian and Shi, Kun and Gan, Xiafan and Li, Guorong and Huang, Zhi},
	month = mar,
	year = {2020},
	pages = {119297},
}

@article{gkerekos_novel_2020,
	title = {A novel, data-driven heuristic framework for vessel weather routing},
	volume = {197},
	issn = {00298018},
	doi = {10.1016/j.oceaneng.2019.106887},
	language = {en},
	urldate = {2024-03-06},
	journal = {Ocean Engineering},
	author = {Gkerekos, Christos and Lazakis, Iraklis},
	month = feb,
	year = {2020},
	pages = {106887},
}

@article{uyanik_machine_2020,
	title = {Machine learning approach to ship fuel consumption: {A} case of container vessel},
	volume = {84},
	issn = {13619209},
	shorttitle = {Machine learning approach to ship fuel consumption},
	doi = {10.1016/j.trd.2020.102389},
	language = {en},
	urldate = {2024-03-06},
	journal = {Transportation Research Part D: Transport and Environment},
	author = {Uyan{\i}k, Tayfun and Karatuğ, Çağlar and Arslano{\u g}lu, Yasin},
	month = jul,
	year = {2020},
	pages = {102389},
}

@article{agand_fuel_2023,
	title = {Fuel consumption prediction for a passenger ferry using machine learning and in-service data: {A} comparative study},
	volume = {284},
	issn = {00298018},
	shorttitle = {Fuel consumption prediction for a passenger ferry using machine learning and in-service data},
	doi = {10.1016/j.oceaneng.2023.115271},
	language = {en},
	urldate = {2024-03-06},
	journal = {Ocean Engineering},
	author = {Agand, Pedram and Kennedy, Allison and Harris, Trevor and Bae, Chanwoo and Chen, Mo and Park, Edward J.},
	month = sep,
	year = {2023},
	pages = {115271},
}

@article{du_data_2022,
	title = {Data fusion and machine learning for ship fuel efficiency modeling: {Part} {II} – {Voyage} report data, {AIS} data and meteorological data},
	volume = {2},
	issn = {27724247},
	shorttitle = {Data fusion and machine learning for ship fuel efficiency modeling},
	doi = {10.1016/j.commtr.2022.100073},
	language = {en},
	urldate = {2024-03-06},
	journal = {Communications in Transportation Research},
	author = {Du, Yuquan and Chen, Yanyu and Li, Xiaohe and Schönborn, Alessandro and Sun, Zhuo},
	month = dec,
	year = {2022},
	pages = {100073},
}

@article{li_data_2022,
	title = {Data fusion and machine learning for ship fuel efficiency modeling: {Part} {I} – {Voyage} report data and meteorological data},
	volume = {2},
	issn = {27724247},
	shorttitle = {Data fusion and machine learning for ship fuel efficiency modeling},
	doi = {10.1016/j.commtr.2022.100074},
	language = {en},
	urldate = {2024-03-06},
	journal = {Communications in Transportation Research},
	author = {Li, Xiaohe and Du, Yuquan and Chen, Yanyu and Nguyen, Son and Zhang, Wei and Schönborn, Alessandro and Sun, Zhuo},
	month = dec,
	year = {2022},
	pages = {100074},
}

@article{du_data_2022-1,
	title = {Data fusion and machine learning for ship fuel efficiency modeling: {Part} {III} – {Sensor} data and meteorological data},
	volume = {2},
	issn = {27724247},
	shorttitle = {Data fusion and machine learning for ship fuel efficiency modeling},
	doi = {10.1016/j.commtr.2022.100072},
	language = {en},
	urldate = {2024-03-06},
	journal = {Communications in Transportation Research},
	author = {Du, Yuquan and Chen, Yanyu and Li, Xiaohe and Schönborn, Alessandro and Sun, Zhuo},
	month = dec,
	year = {2022},
	pages = {100072},
}

@article{hu_two-step_2022,
	title = {A two-step strategy for fuel consumption prediction and optimization of ocean-going ships},
	volume = {249},
	issn = {00298018},
	doi = {10.1016/j.oceaneng.2022.110904},
	language = {en},
	urldate = {2024-03-06},
	journal = {Ocean Engineering},
	author = {Hu, Zhihui and Zhou, Tianrui and Zhen, Rong and Jin, Yongxing and Li, Xiaohe and Osman, Mohd Tarmizi},
	month = apr,
	year = {2022},
	pages = {110904},
}

@article{du_two-phase_2019,
	title = {Two-phase optimal solutions for ship speed and trim optimization over a voyage using voyage report data},
	volume = {122},
	issn = {01912615},
	doi = {10.1016/j.trb.2019.02.004},
	language = {en},
	urldate = {2024-03-07},
	journal = {Transportation Research Part B: Methodological},
	author = {Du, Yuquan and Meng, Qiang and Wang, Shuaian and Kuang, Haibo},
	month = apr,
	year = {2019},
	pages = {88--114},
}

@article{fan_review_2022,
	title = {A review of ship fuel consumption models},
	volume = {264},
	issn = {00298018},
	doi = {10.1016/j.oceaneng.2022.112405},
	language = {en},
	urldate = {2024-03-11},
	journal = {Ocean Engineering},
	author = {Fan, Ailong and Yang, Jian and Yang, Liu and Wu, Da and Vladimir, Nikola},
	month = nov,
	year = {2022},
	pages = {112405},
}

@article{hu_novel_2021,
	title = {A {Novel} {Hybrid} {Fuel} {Consumption} {Prediction} {Model} for {Ocean}-{Going} {Container} {Ships} {Based} on {Sensor} {Data}},
	volume = {9},
	issn = {2077-1312},
	doi = {10.3390/jmse9040449},
	language = {en},
	number = {4},
	urldate = {2024-03-11},
	journal = {Journal of Marine Science and Engineering},
	author = {Hu, Zhihui and Zhou, Tianrui and Osman, Mohd Tarmizi and Li, Xiaohe and Jin, Yongxin and Zhen, Rong},
	month = apr,
	year = {2021},
	pages = {449},
}

@article{szlapczynska_preference-based_2019,
	title = {Preference-based evolutionary multi-objective optimization in ship weather routing},
	volume = {84},
	issn = {15684946},
	doi = {10.1016/j.asoc.2019.105742},
	language = {en},
	urldate = {2024-03-14},
	journal = {Applied Soft Computing},
	author = {Szlapczynska, Joanna and Szlapczynski, Rafal},
	month = nov,
	year = {2019},
	pages = {105742},
}

@article{tu_optimum_2023,
	title = {Optimum trim prediction for container ships based on machine learning},
	volume = {277},
	issn = {00298018},
	doi = {10.1016/j.oceaneng.2022.111322},
	language = {en},
	urldate = {2024-03-14},
	journal = {Ocean Engineering},
	author = {Tu, Haiwen and Xia, Kai and Zhao, Enjin and Mu, Lin and Sun, Jianglong},
	month = jun,
	year = {2023},
	pages = {111322},
}

@article{coraddu_vessels_2017,
	title = {Vessels fuel consumption forecast and trim optimisation: {A} data analytics perspective},
	volume = {130},
	issn = {00298018},
	shorttitle = {Vessels fuel consumption forecast and trim optimisation},
	doi = {10.1016/j.oceaneng.2016.11.058},
	language = {en},
	urldate = {2024-03-18},
	journal = {Ocean Engineering},
	author = {Coraddu, Andrea and Oneto, Luca and Baldi, Francesco and Anguita, Davide},
	month = jan,
	year = {2017},
	pages = {351--370},
}

@article{fan_novel_2020,
	title = {A novel ship energy efficiency model considering random environmental parameters},
	volume = {19},
	issn = {2046-4177, 2056-8487},
	doi = {10.1080/20464177.2018.1546644},
	language = {en},
	number = {4},
	urldate = {2024-03-18},
	journal = {Journal of Marine Engineering \& Technology},
	author = {Fan, Ailong and Yan, Xinping and Bucknall, Richard and Yin, Qizhi and Ji, Sheng and Liu, Yuanchang and Song, Rui and Chen, Xiaping},
	month = nov,
	year = {2020},
	pages = {215--228},
}

@article{yang_genetic_2019,
	title = {A genetic algorithm-based grey-box model for ship fuel consumption prediction towards sustainable shipping},
	issn = {0254-5330, 1572-9338},
	doi = {10.1007/s10479-019-03183-5},
	language = {en},
	urldate = {2024-03-19},
	journal = {Annals of Operations Research},
	author = {Yang, Liqian and Chen, Gang and Rytter, Niels Gorm Malý and Zhao, Jinlou and Yang, Dong},
	month = mar,
	year = {2019},
}

@article{meng_shipping_2016,
	title = {Shipping log data based container ship fuel efficiency modeling},
	volume = {83},
	issn = {01912615},
	doi = {10.1016/j.trb.2015.11.007},
	language = {en},
	urldate = {2024-03-19},
	journal = {Transportation Research Part B: Methodological},
	author = {Meng, Qiang and Du, Yuquan and Wang, Yadong},
	month = jan,
	year = {2016},
	pages = {207--229},
}

@article{wang_comprehensive_2022,
	title = {A comprehensive review on the prediction of ship energy consumption and pollution gas emissions},
	volume = {266},
	issn = {00298018},
	doi = {10.1016/j.oceaneng.2022.112826},
	language = {en},
	urldate = {2024-03-25},
	journal = {Ocean Engineering},
	author = {Wang, Kai and Wang, Jianhang and Huang, Lianzhong and Yuan, Yupeng and Wu, Guitao and Xing, Hui and Wang, Zhongyi and Wang, Zhuang and Jiang, Xiaoli},
	month = dec,
	year = {2022},
	pages = {112826},
}

@article{guo_combined_2022,
	title = {Combined machine learning and physics-based models for estimating fuel consumption of cargo ships},
	volume = {255},
	issn = {00298018},
	doi = {10.1016/j.oceaneng.2022.111435},
	language = {en},
	urldate = {2024-03-26},
	journal = {Ocean Engineering},
	author = {Guo, Bingjie and Liang, Qin and Tvete, Hans Anton and Brinks, Hendrik and Vanem, Erik},
	month = jul,
	year = {2022},
	pages = {111435},
}

@article{chen_short-term_2024,
	title = {Short-term forecasting for ship fuel consumption based on deep learning},
	volume = {301},
	issn = {00298018},
	doi = {10.1016/j.oceaneng.2024.117398},
	language = {en},
	urldate = {2024-04-04},
	journal = {Ocean Engineering},
	author = {Chen, Yumei and Sun, Baozhi and Xie, Xianwei and Li, Xiaohe and Li, Yanjun and Zhao, Yuhao},
	month = jun,
	year = {2024},
	pages = {117398},
}

@article{xie_fuel_2023,
	title = {Fuel {Consumption} {Prediction} {Models} {Based} on {Machine} {Learning} and {Mathematical} {Methods}},
	volume = {11},
	issn = {2077-1312},
	doi = {10.3390/jmse11040738},
	language = {en},
	number = {4},
	urldate = {2024-04-04},
	journal = {Journal of Marine Science and Engineering},
	author = {Xie, Xianwei and Sun, Baozhi and Li, Xiaohe and Olsson, Tobias and Maleki, Neda and Ahlgren, Fredrik},
	month = mar,
	year = {2023},
	pages = {738},
}

@incollection{karakostas_enhanced_2024,
	title = {Enhanced and {Holistic} {Voyage} {Planning} {Using} {Digital} {Twins}:},
	copyright = {http://creativecommons.org/licenses/by/3.0/deed.en\_US},
	isbn = {978-1-66849-848-4 978-1-66849-849-1},
	shorttitle = {Enhanced and {Holistic} {Voyage} {Planning} {Using} {Digital} {Twins}},
	language = {en},
	urldate = {2024-04-25},
	booktitle = {Advances in {Logistics}, {Operations}, and {Management} {Science}},
	publisher = {IGI Global},
	author = {Kaklis, Dimitris and Antonopoulos, Antonis},
	editor = {Karakostas, Bill and Katsoulakos, Takis},
	month = apr,
	year = {2024},
	doi = {10.4018/978-1-6684-9848-4.ch006},
	pages = {112--136},
}

@article{kim_modelling_2023,
	title = {Modelling of ship resistance and power consumption for the global fleet: {The} {MariTEAM} model},
	volume = {281},
	issn = {00298018},
	shorttitle = {Modelling of ship resistance and power consumption for the global fleet},
	doi = {10.1016/j.oceaneng.2023.114758},
	language = {en},
	urldate = {2024-04-29},
	journal = {Ocean Engineering},
	author = {Kim, Young-Rong and Steen, Sverre and Kramel, Diogo and Muri, Helene and Strømman, Anders Hammer},
	month = aug,
	year = {2023},
	pages = {114758},
}

@phdthesis{zhang_physics-infused_2021,
	title = {Physics-infused {Hybrid} {Machine} {Learning} {Models} and {Their} {Applications}},
	language = {en},
	school = {The State University of New York},
	author = {Zhang, Zhibo},
	year = {2021},
}

@article{khan_benefits_2022,
	title = {The benefits of co-evolutionary {Genetic} {Algorithms} in voyage optimisation},
	volume = {245},
	issn = {00298018},
	doi = {10.1016/j.oceaneng.2021.110261},
	language = {en},
	urldate = {2024-04-30},
	journal = {Ocean Engineering},
	author = {Khan, Saima and Grudniewski, Przemyslaw and Muhammad, Yousaf Shad and Sobey, Adam J.},
	month = feb,
	year = {2022},
	pages = {110261},
}

@article{kaneko_hybrid_2023,
	title = {Hybrid physics-based and machine learning model with interpretability and uncertainty for real-time estimation of unmeasurable parts},
	volume = {284},
	issn = {00298018},
	doi = {10.1016/j.oceaneng.2023.115267},
	language = {en},
	urldate = {2024-05-23},
	journal = {Ocean Engineering},
	author = {Kaneko, Tatsuya and Wada, Ryota and Ozaki, Masahiko and Inoue, Tomoya},
	month = sep,
	year = {2023},
	pages = {115267},
}

@article{kuhlemann_genetic_2020,
	title = {A genetic algorithm for finding realistic sea routes considering the weather},
	volume = {26},
	issn = {1381-1231, 1572-9397},
	doi = {10.1007/s10732-020-09449-7},
	language = {en},
	number = {6},
	urldate = {2024-05-27},
	journal = {Journal of Heuristics},
	author = {Kuhlemann, Stefan and Tierney, Kevin},
	month = dec,
	year = {2020},
	pages = {801--825},
}

@book{thuerey_physics-based_2022,
	title = {Physics-based {Deep} {Learning}},
	urldate = {2024-05-28},
	publisher = {arXiv},
	author = {Thuerey, Nils and Holl, Philipp and Mueller, Maximilian and Schnell, Patrick and Trost, Felix and Um, Kiwon},
	month = apr,
	year = {2022},
	note = {arXiv:2109.05237 [physics]},
}

@article{hajli_fuel_2024,
	title = {A fuel consumption prediction model for ships based on historical voyages and meteorological data},
	issn = {2046-4177, 2056-8487},
	doi = {10.1080/20464177.2024.2371192},
	language = {en},
	urldate = {2024-07-17},
	journal = {Journal of Marine Engineering \& Technology},
	author = {Hajli, Kaoutar and Rönnqvist, Mikael and Dadouchi, Camélia and Audy, Jean-François and Cordeau, Jean-François and Warya, Gurjeet and Ngo, Trung},
	month = jul,
	year = {2024},
	pages = {1--12},
}

@article{bayraktar_marine_2024,
	title = {Marine vessel energy efficiency performance prediction based on daily reported noon reports},
	volume = {19},
	issn = {1744-5302, 1754-212X},
	doi = {10.1080/17445302.2023.2214490},
	language = {en},
	number = {6},
	urldate = {2024-07-17},
	journal = {Ships and Offshore Structures},
	author = {Bayraktar, Murat and Sokukcu, Mustafa},
	month = jun,
	year = {2024},
	pages = {831--840},
}

@article{ilias_multitask_2023,
	title = {A {Multitask} {Learning} {Framework} for {Predicting} {Ship} {Fuel} {Oil} {Consumption}},
	volume = {11},
	copyright = {https://creativecommons.org/licenses/by-nc-nd/4.0/},
	issn = {2169-3536},
	doi = {10.1109/ACCESS.2023.3335905},
	language = {en},
	urldate = {2024-07-19},
	journal = {IEEE Access},
	author = {Ilias, Loukas and Kapsalis, Panagiotis and Mouzakitis, Spiros and Askounis, Dimitris},
	year = {2023},
	pages = {132576--132589},
}

@inproceedings{kaklis_online_2022,
	address = {Paphos, Cyprus},
	title = {Online {Training} for {Fuel} {Oil} {Consumption} {Estimation}: {A} {Data} {Driven} {Approach}},
	copyright = {https://doi.org/10.15223/policy-029},
	isbn = {978-1-66545-176-5},
	shorttitle = {Online {Training} for {Fuel} {Oil} {Consumption} {Estimation}},
	doi = {10.1109/MDM55031.2022.00088},
	language = {en},
	urldate = {2024-07-22},
	booktitle = {2022 23rd {IEEE} {International} {Conference} on {Mobile} {Data} {Management} ({MDM})},
	publisher = {IEEE},
	author = {Kaklis, Dimitrios and Varlamis, Iraklis and Giannakopoulos, George and Spyropoulos, Constantine and Varelas, Takis J.},
	month = jun,
	year = {2022},
	pages = {394--400},
}

@article{yuksel_comparative_2023,
	title = {Comparative study of machine learning techniques to predict fuel consumption of a marine diesel engine},
	volume = {286},
	issn = {00298018},
	doi = {10.1016/j.oceaneng.2023.115505},
	language = {en},
	urldate = {2024-07-22},
	journal = {Ocean Engineering},
	author = {Yuksel, Onur and Bayraktar, Murat and Sokukcu, Mustafa},
	month = oct,
	year = {2023},
	pages = {115505},
}

@article{nguyen_application-oriented_2023,
	title = {An application-oriented testing regime and multi-ship predictive modeling for vessel fuel consumption prediction},
	volume = {177},
	issn = {13665545},
	doi = {10.1016/j.tre.2023.103261},
	language = {en},
	urldate = {2024-07-23},
	journal = {Transportation Research Part E: Logistics and Transportation Review},
	author = {Nguyen, Son and Fu, Xiuju and Ogawa, Daichi and Zheng, Qin},
	month = sep,
	year = {2023},
	pages = {103261},
}

@article{barreiro_review_2022,
	title = {Review of ship energy efficiency},
	volume = {257},
	issn = {00298018},
	doi = {10.1016/j.oceaneng.2022.111594},
	language = {en},
	urldate = {2024-07-23},
	journal = {Ocean Engineering},
	author = {Barreiro, Julio and Zaragoza, Sonia and Diaz-Casas, Vicente},
	month = aug,
	year = {2022},
	pages = {111594},
}

@article{wang2023data,
  title={Data-driven methods for detection of abnormal ship behavior: Progress and trends},
  author={Wang, Yukuan and Liu, Jingxian and Liu, Ryan Wen and Liu, Yang and Yuan, Zhi},
  journal={Ocean Engineering},
  volume={271},
  pages={113673},
  year={2023},
  publisher={Elsevier}
}

@article{FERRARI2023100985,
title = {The impact of rising maritime transport costs on international trade: Estimation using a multi-region general equilibrium model},
journal = {Transportation Research Interdisciplinary Perspectives},
volume = {22},
pages = {100985},
year = {2023},
issn = {2590-1982},
author = {Emanuele Ferrari and Panayotis Christidis and Paolo Bolsi}
}

@article{handayani2023navigating,
  title={Navigating energy efficiency: A multifaceted interpretability of fuel oil consumption prediction in cargo container vessel considering the operational and environmental factors},
  author={Handayani, Melia Putri and Kim, Hyunju and Lee, Sangbong and Lee, Jihwan},
  journal={Journal of Marine Science and Engineering},
  volume={11},
  number={11},
  pages={2165},
  year={2023},
  publisher={MDPI}
}

@article{handayani2025predictive,
  title={Predictive analysis of fuel oil consumption in vessels: interpretable modelling with emphasis on load conditions},
  author={Handayani, Melia Putri and Kim, Donghyun and Lee, Sangbong and Lee, Jihwan},
  journal={Maritime Policy \& Management},
  pages={1--24},
  year={2025},
  publisher={Taylor \& Francis}
}

@article{li2024dapnet,
  title={DAPNet: A Dual-Attention Parallel Network for the Prediction of Ship Fuel Consumption Based on Multi-Source Data},
  author={Li, Xinyu and Zuo, Yi and Jiang, Junhao},
  journal={Journal of Marine Science and Engineering},
  volume={12},
  number={11},
  pages={1945},
  year={2024},
  publisher={MDPI}
}

\end{document}